\documentclass[acmsmall]{acmart}
\newtheorem*{Corollary}{Corollary}
\usepackage{multirow}
\usepackage{multicol}
\usepackage{colortbl}
\usepackage{subfigure}
\usepackage{algorithm}
\usepackage{algpseudocode}

\usepackage{enumitem}

\AtBeginDocument{%
  \providecommand\BibTeX{{%
    \normalfont B\kern-0.5em{\scshape i\kern-0.25em b}\kern-0.8em\TeX}}}

\setcopyright{acmlicensed}
\copyrightyear{2018}
\acmYear{2018}
\acmDOI{XXXXXXX.XXXXXXX}

\acmConference[Conference acronym 'XX]{Make sure to enter the correct
  conference title from your rights confirmation emai}{June 03--05,
  2018}{Woodstock, NY}
\acmBooktitle{Woodstock '18: ACM Symposium on Neural Gaze Detection,
 June 03--05, 2018, Woodstock, NY} 
\acmISBN{978-1-4503-XXXX-X/18/06}

\begin{document}


\title{Symmetric Graph Contrastive Learning against Noisy Views\\ for Recommendation}

\author{Chu Zhao}
\email{chuzhao@stumail.neu.edu.cn}
\affiliation{%
  \institution{Software College, Northeastern University}
  \city{Shenyang}
  \country{China}}

\author{Enneng Yang}
\email{ennengyang@stumail.neu.edu.cn}
\affiliation{%
  \institution{Software College, Northeastern University}
  \city{Shenyang}
  \country{China}}

\author{Yuliang Liang}
\email{liangyuliang@stumail.neu.edu.cn}
\affiliation{%
  \institution{Software College, Northeastern University}
  \city{Shenyang}
  \country{China}}

\author{Jianzhe Zhao}
\email{zhaojz@swc.neu.edu.cn}
\authornote{Corresponding authors.}
\affiliation{%
  \institution{Software College, Northeastern University}
  \city{Shenyang}
  \country{China}}

\author{Guibing Guo}
\email{guogb@swc.neu.edu.cn}
\authornotemark[1]
\affiliation{%
  \institution{Software College, Northeastern University}
  \city{Shenyang}
  \country{China}}

\author{Xingwei Wang}
\email{wangxw@mail.neu.edu.cn}
\affiliation{%
  \institution{School of Computer Science and Engineering, Northeastern University}
  \city{Shenyang}
  \country{China}}

\renewcommand{\shortauthors}{Zhao and Yang, et al.}

\begin{abstract}
Graph Contrastive Learning (GCL) leverages data augmentation techniques to produce contrasting views, enhancing the accuracy of recommendation systems through learning the consistency between contrastive views.
However, existing augmentation methods, such as directly perturbing interaction graph (e.g., node/edge dropout), may interfere with the original connections and generate poor contrasting views, resulting in sub-optimal performance. In this paper, we define the views that share only a small amount of information with the original graph due to poor data augmentation as noisy views (i.e., the last 20\% of the views with a cosine similarity value less than 0.1 to the original view). We demonstrate through detailed experiments that noisy views will significantly degrade recommendation performance. Further, we propose a model-agnostic Symmetric Graph Contrastive Learning (SGCL) method with theoretical guarantees to address this issue. Specifically, we introduce symmetry theory into graph contrastive learning, based on which we propose a symmetric form and contrast loss resistant to noisy interference. We provide theoretical proof that our proposed SGCL method has a high tolerance to noisy views. Further demonstration is given by conducting extensive experiments on three real-world datasets. The experimental results demonstrate that our approach substantially increases recommendation accuracy, with relative improvements reaching as high as 12.25\% over nine other competing models. These results highlight the efficacy of our method. The code is available at \url{https://github.com/user683/SGCL}. 

\end{abstract}



\maketitle

\section{Introduction}
Graph Convolutional Networks  (GCNs) have recently garnered extensive attention in the recommendation field. Benefiting from the capability to learn sophisticated representations, GCN-based recommendation systems can model user preferences from user-item interactions \cite{wu2022graph,wu2020comprehensive}. They generally transform the interactions into a graph and take them as input. Then, they utilize GCNs to distribute the embeddings across neighboring nodes within the user-item interaction graph, thereby exploring higher-order connectivity. Ultimately, models use the inner product to calculate the user and item embeddings and predict user preference for the item.
Several advanced GCN-based collaborative filtering frameworks have been proposed, including LightGCN \cite{he2020lightgcn}, NGCF \cite{wang2019neural}, and DGCF \cite{wang2020disentangled}.

\begin{figure}[t]
	 \centering
	\begin{minipage}{0.8\linewidth}
\centerline{\includegraphics[width=\textwidth]{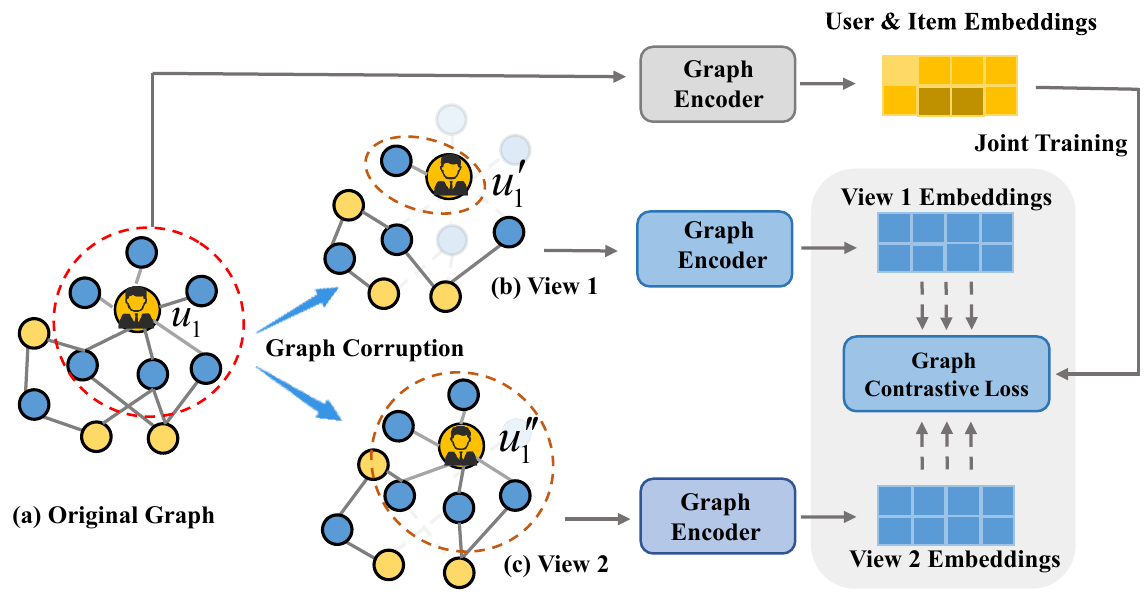}}
	\end{minipage}
	\caption{Overall framework of graph contrastive learning for collaborative filtering. It is worth mentioning that view 1 and view 2 are two generated contrastive views of $u_1$ (i.e., original graph). We observe that the structure of $u'_1$ is different from the original and $u''_1$, which hints that graph augmentation may introduce noisy views and thus degrade the final recommendation performance}.
	\label{fig1}
\end{figure}

While these models have demonstrated success, their reliance on high-quality labels (i.e., supervised signals) presents a drawback, particularly in real-world scenarios; obtaining a large amount of labeled data is impractical due to constraints on resources and privacy protection regulations. User behaviors often exhibit noise and inconsistency. Such a situation will lead to suboptimal representations and hinder the improvement of recommendation performance. Research indicates that self-supervised learning can harness additional signals from unlabeled data and utilize contrastive learning to enhance representation learning \cite{jaiswal2020survey,liu2021self,schiappa2023self,liu2022graph, jing2023contrastive, fang2020deep}. Motivated by these works, researchers have designed a series of self-supervised graph contrastive learning recommendation frameworks \cite{yu2023self,ren2023sslrec}. These methods employ various cleverly designed augmentation operations, such as node/edge dropout, to generate contrasting views (i.e., manually generating contrastive views), and we displayed the contrastive learning general framework in Figure \ref{fig1}. Subsequently, the contrastive loss function, such as InfoNCE, is used to ensure the consistency of embeddings by increasing the mutual information among positive sample pairs while reducing it among negative pairs \cite{raghunathan2018semidefinite, oord2018representation}. 

\begin{figure}[t]

	 \centering
	\begin{minipage}{0.31\linewidth}
\centerline{\includegraphics[width=\textwidth]{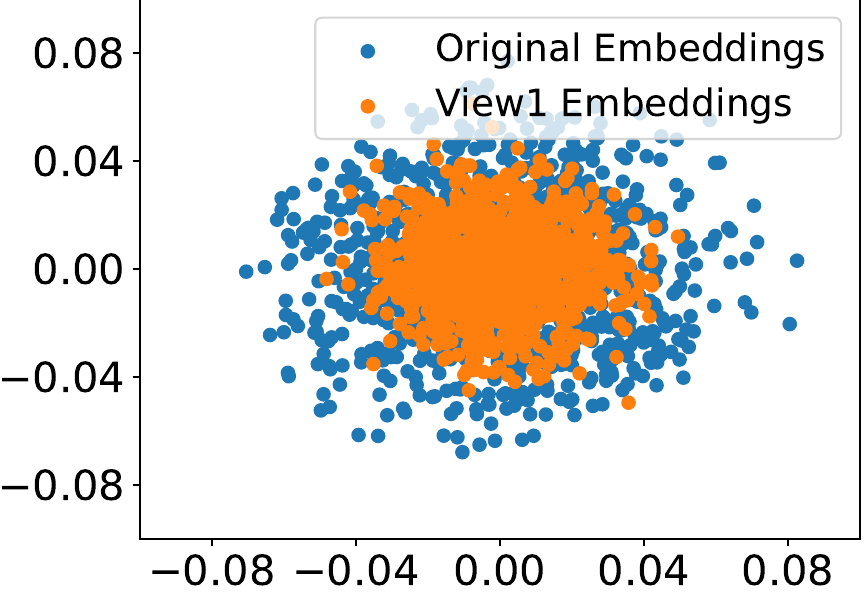}}
 \centerline{\;\;\;\;\;\;\;(a)}
	\end{minipage}
 \begin{minipage}{0.31\linewidth}
  \centerline{\includegraphics[width=\textwidth]{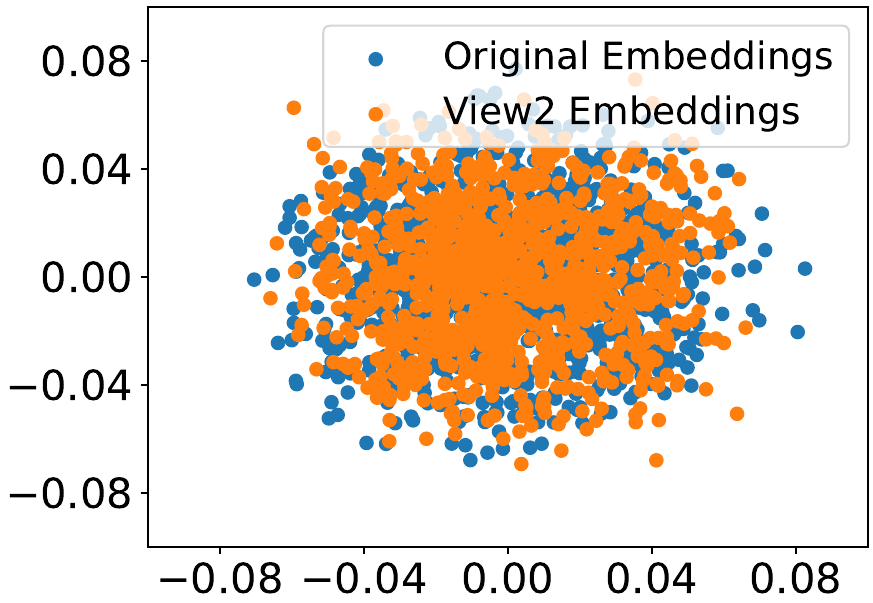}}
		\centerline{\;\;\;\;\;\;\;(b)}
	\end{minipage}
 \begin{minipage}{0.33\linewidth}
 \vspace{1pt}
\centerline{\includegraphics[width=\textwidth]{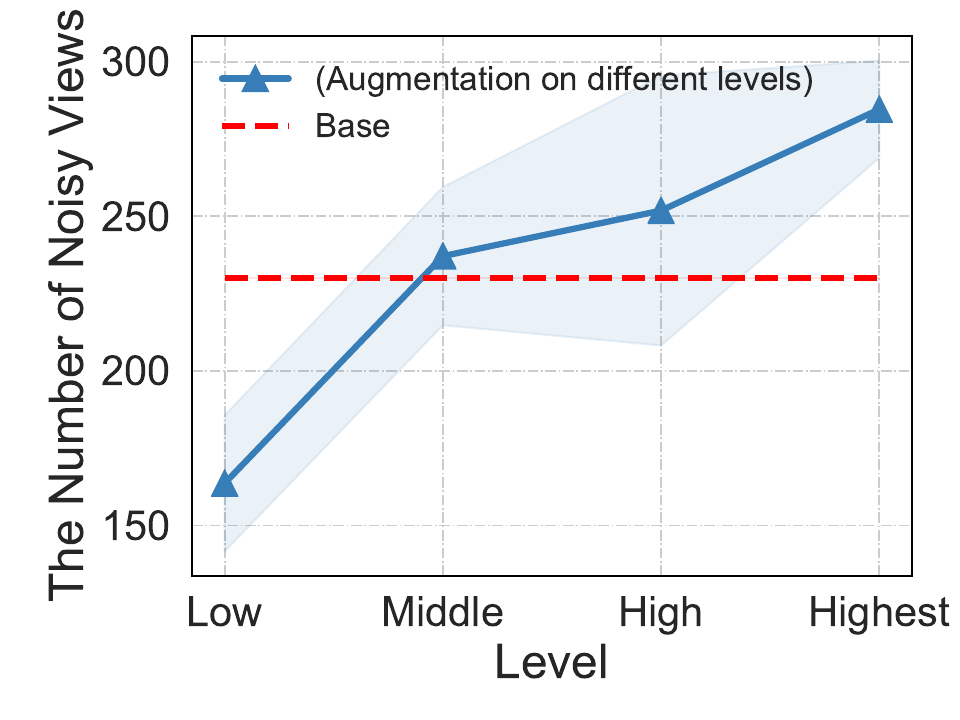}}
    \vspace{-10pt}
		\centerline{\hspace{20pt}(c)}
	\end{minipage}
	\caption{Part (a) is the visualization of the embeddings of \emph{View 1} and the visualization of the embeddings of \emph{Original}. Part (b) denotes the visualization of the embeddings of \emph{View 2} and the visualization of the embeddings of \emph{Original}. Part (c) shows the number of noisy views in different level groups.}
	\label{fig2}
\end{figure}

However, we argue that such heuristic-based data augmentation operations may remove crucial nodes or edges from the graph, resulting in less shared learnable information between the generated and original views. In this work, we define the generated view that shares only a small amount of information with the original view in the process of graph data augmentation as \emph{noisy views}. Take Figure \ref{fig1} as an example; two contrasting views (i.e., \emph{View 1} and \emph{View 2}) are generated for user $u_1$ through edge dropout.
It can be observed that there is a significant difference in the graph structure between the generated view (\emph{View 1}) and the original view. In Figure \ref{fig2}, we further visualize the original and generated views' embeddings. 
From the visualizations in Figure \ref{fig2}, it is evident that \emph{View 1} preserves less original semantic information compared with \emph{View 2}. In this scenario, aligning these two contrasting views is inappropriate, as it would result in suboptimal representations and detrimentally affect recommendation performance.
Recently, some studies have devised adaptive CL methods to avoid directly perturbing the graph  (and thus not generating contrastive views), including AutoCF \cite{xia2023automated}, AdaGCL \cite{Jiang2023adagcl}, and HCCF \cite{xia2022hypergraph}.
The above methods directly show their motivation that manually performing data augmentation may disturb the original
connectivity, leading to the learning of suboptimal representations.
However, these methods overlook the effects of corruption on graph and how it impacts recommendation performance, which may not adequately justify their motivation.

Motivated by these observations, we first design comprehensive experiments to explore such manual graph augmentations and how they impact the accuracy of the recommendation. 
In Section \ref{section 2.1}, we offer an in-depth explanation of our experimental framework, outcomes, and the rationale for our study. The data presented in Table \ref{tab01_1} illustrate a drop in performance of the recommendation model following the removal of edges deemed highly important.
Figure \ref{fig2} (c) indicates that the noisy view ratio increases as the importance of deleted edges increases. Hence, based on the aforementioned observations, we can conclude the following: 1) During the manual graph enhancement process, a noisy view will be generated if important edges are removed. 2) Recommendation accuracy will be decreased if noisy views are generated and then ignored by these perturbations during the training process.

Considering these constraints, it becomes crucial to develop theoretically solid, graph contrastive learning techniques that are specifically designed to address noisy views. Inspired by symmetric loss function theory, we intend to develop a novel contrastive loss approach rooted in symmetry theory to bolster the model's resilience against noisy views. Existing work \cite{ghosh2017robust}
indicates that the loss function is noise-tolerant if the minimizer of risk (under that loss function) with noisy labels would be the same as those with noise-free labels.
Therefore, providing a symmetric loss can reduce the model’s loss in a noisy environment, enhancing accuracy performance. This theoretical concept is known as Noise-Tolerant Symmetric Loss \cite{ghosh2017robust}. However, applying this appealing technique to GCL-based recommendation systems presents some challenges:
(i) The existing graph contrastive learning loss functions, such as infoNCE, do not satisfy the symmetry condition. Therefore, it is essential to forge a link between symmetric loss and contrastive learning, converting it into a format that meets the symmetry condition. (ii) Additionally, the proposed new contrastive learning loss should be demonstrated to satisfy the property of being a lower bound for mutual information.

To bridge these gaps, in this study, we design a Symmetric Graph Contrastive Learning (SGCL) method aimed at enhancing recommendation performance. Unlike these adaptive recommendation methods, our study first defines the noisy view and introduces a symmetric contrastive loss (SCL) that satisfies the symmetric condition against noisy views for the recommendation. With the proposed SGCL, we first introduce the symmetric form of graph contrastive learning and then propose the symmetric loss for graph contrastive learning, providing theoretical proof of the capability against noisy views. Simultaneously, we show that the SGCL function serves as a lower bound for the Wasserstein Dependency Measure (WDM), that is, it satisfies the property of being a lower bound of mutual information. It's noteworthy that SGCL does not prevent the generation of noisy views but rather enhances the model's tolerance to them, making it easily adaptable to other graph contrastive learning methods.
In summary, the primary contributions of this paper can be outlined as follows:
 \begin{itemize}[leftmargin=*]
     \item \textbf{General Aspect}. We conduct experiments to explore the impact of manual data augmentation on the recommendation performance based on the user-item interaction graph. Our findings reveal that such augmentation may disturb the inherent connectivity of the graph, leading to the emergence of noisy views. Consequently, these noisy views contribute to a decrease in the performance of the recommendation system.
     \item \textbf{Methodology}. We introduce SGCL, a \underline{S}ymmetric \underline{G}raph \underline{C}ontrastive \underline{L}earning approach that is independent of specific models. SGCL utilizes a novel contrastive loss to address noisy views, thereby enhancing the performance of recommendation systems.
     In addition, we provided theoretical proof of SGCL against noisy views.
     \item \textbf{Experimental Findings}. Our experiments on three datasets reveal that SGCL consistently outperforms a range of baseline methods. In detail, our SGCL achieves stronger robustness when facing the challenge of noisy and sparse data, and the maximum metric improvement rate reaches 12.25\% compared to the baselines.
 \end{itemize}

\section{Motivation}
In Section 2, we first visualized the embeddings of two contrastive views and the original graph. Then, we conducted experiments to explore the impact of noisy views on recommender systems. 
\subsection{Noisy Views in Data Augmentation}
\label{section 2.1}
 We direct the perturbation of the interaction graph to generate noisy views in this section.
 We initially and respectively extract 100,000 user-item interactions from the Yelp2020 \footnote{\footnotesize \url{https://www.yelp.com/dataset}} and Amazon-CD \footnote{\footnotesize \url{https://nijianmo.github.io/amazon/index.html}} datasets to generate a bipartite graph. Subsequently, we calculate the importance of each edge/node in the interaction graph with
 the Betweenness Centrality \cite{brandes2001faster}. A larger value means that the corresponding edge/node is more important. The interaction edges/nodes are classified into four levels based on their importance. Specifically, we first sort the data in descending order of importance. The top 1\% of the most important edges are categorized as $Highest$, the data from the top 2\% to 4\% is labeled as $High$, the data from the top 5\% to 10\% is assigned to $Middle$, and the remaining 90\% of the data is categorized as $Low$. We sequentially leverage the random edge/node dropout at different levels to construct contrastive views with a dropout ratio of 0.1.
Simultaneously, the cosine similarity is leveraged to measure the similarity between the generated and original views. 
Without loss of generality, we define the last 20\% of data with values less than 0.1 as the \textbf{noisy view}.

Figure \ref{fig2} (a) and Figure \ref{fig2} (b) visualize the embedding of noisy view \emph{View 1}, the embedding of \emph{View 2}, and the embedding of original view, respectively. We can find that the embeddings of \emph{View 1} only partially overlap with the embeddings of the original \emph{View}. In contrast, the embedding of \emph{View 2} largely overlaps with the embeddings of the original \emph{View}. The above observations indicate that directly perturbing interaction graph will generate contrastive views with less shared information.
Figure \ref{fig2} (c) reports the number of noisy views when generating contrastive views at different levels. We can see that when constructing contrastive views in groups with higher edge importance (i.e., removing more critical edges), the resulting quantity of noisy views increases. This indicates that directly perturbing the graph structure to generate contrastive views and removing critical edges is more likely to generate noisy views. 

In conclusion, the experimental findings indicate that disturbances to the graph lead to noisy views, and removing more crucial edges further amplifies the likelihood of creating noisy views. We have the same conclusion when using the node dropout to generate contrastive views.

\begin{table}[t]
\centering
 \caption{Performance comparison in terms of NDCG@20 when \textbf{crucial edges} at varying importance levels are removed in different datasets. The column \emph{Base} denotes the variant of random data augmentation without distinguishing edge weights. It is used as the basic baseline.}
 \renewcommand{\arraystretch}{1} 
 \setlength{\tabcolsep}{1mm}{ 
    \begin{tabular}{c|cccccc}
     \hline    
      \rowcolor{gray!20} Dataset & Method & Base & Low & Middle  & High & Highest \\
    \hline
    \multirow{4}{*}{Amazon-CD} & SGL & 0.0452 & 0.0443 & 0.0435  & 0.0420 & 0.0400 \\
     & & &-1.99\% & -3.76\% & -7.08\% & -11.50\% \\
    & KGCL& 0.0501 & 0.0476  & 0.0451 &  0.0439 & 0.0411 \\
    & & &-4.99\% & -9.98\% & -12.38\% & -17.96\% \\
    \hline
    \multirow{4}{*}{Yelp2020}& SGL & 0.0831  & 0.0822 &
    0.0822 & 0.0815 & 0.0802 \\
    & & &-1.08\% & -1.08\% & -1.93\% & -3.49\% \\
    & KGCL & 0.0873 & 0.0856 &0.0854 & 0.0810 & 0.0795 \\
    & & &-1.95\% & -2.18\% & -7.22\% & -8.93\% \\
    \hline
    \end{tabular}}
    \label{tab01_1}
\end{table}

\begin{table}[t]
\centering
\vspace{-10pt}
 \caption{Performance comparison in terms of NDCG@20 when \textbf{crucial nodes} at varying importance levels are removed in different datasets. The column \emph{Base} denotes the variant of random data augmentation without distinguishing node weights. It is used as the basic baseline.}
 \renewcommand{\arraystretch}{1} 
 \setlength{\tabcolsep}{1mm}{ 
    \begin{tabular}{c|cccccc}
     \hline    
      \rowcolor{gray!20} Dataset & Method & Base & Low & Middle & High & Highest \\
    \hline
    \multirow{4}{*}{Amazon-CD} & SGL & 0.0450 & 0.0440 & 0.0431  & 0.0418 & 0.0398 \\
     & & &-2.22\% & -4.22\% & -7.11\% & -11.56\% \\
    & KGCL& 0.0491 & 0.0476  & 0.0449 &  0.0438 & 0.0410 \\
    & & &-3.05\% & -8.55\% & -10.97\% & -16.50\% \\
    \hline
    \multirow{4}{*}{Yelp2020}& SGL & 0.0829  & 0.0820 &
    0.0811 & 0.0805 & 0.0799 \\
    & & &-1.09\% & -2.17\% & -2.09\% & -3.62\% \\
    & KGCL & 0.0870 & 0.0848 &0.0832 & 0.0806 & 0.0729 \\
    & & &-2.53\% & -4.37\% & -7.36\% & -16.21\% \\
    \hline
    \end{tabular}}
    \label{tab01_2}
\end{table}

\subsection{Investigating the Impact of Noisy Views on Recommendation Accuracy}
In this section, the edge and node dropout are employed to construct contrastive views, respectively, and compare the performance changes of two competing GNN-based recommendation models. 
Specifically, the following two methods are selected for the pilot study: (1) \textbf{SGL} \cite{wu2021self} employs LightGCN \cite{he2020lightgcn} as the backbone and designs a series of structure augmentations to assist representation learning. (2) \textbf{KGCL} \cite{yang2022knowledge} is a contrastive learning paradigm enhanced by a knowledge graph, which mitigates the long-tail and noise issues in the knowledge graph. Our experimenters use the KGCL (KGCL w/o KGA) variant, which does not employ the knowledge-guided augmentation scheme on the bipartite graph. 

We display the experimental results with edge and node dropout in Table \ref{tab01_1} and Table \ref{tab01_2}, respectively. We can see that: (1) On both datasets, SGL and KGCL consistently exhibit lower recommendation accuracy across the four groups than the accuracy on the $Base$. 
Specifically, in Table \ref{tab01_1}, SGL and KGCL show a performance decrease of 11.50\% and 17.96\% in the Amazon-CD dataset, respectively, in the group $Highest$ compared to the performance in the $Base$. In the Yelp2020 dataset, the performance of SGL and KGCL decrease by 3.49\% and 8.93\% in group $Highest$, respectively. From Table \ref{tab01_2}, the metric  NDCG@20 of SGL and KGCL in the $Highest$ group decrease by 11.56\% and 16.50\% compared to the performance in the $Base$ group on the Amazon-CD dataset. On the Yelp2020 dataset, both SGL and KGCL exhibit a similar drop in performance, with NDCG@20 decreasing by 3.62\% and 12.61\%, respectively, compared to the $Base$ group.
(2) With experiments conducted in groups characterized by higher edge importance, both the SGL and KGCL models exhibit a decline in performance. In detail, the recommendation accuracy of both models in group $Highest$ is lower than in the other three groups. The performance of SGL and KGCL in group $High$ is less favorable than group $Middle$ and $Low$. In group $Low$, SGL and KGCL perform better than the other three groups. 

Therefore, we may conclude that the more critical edges or nodes are removed by data augmentation, the more noisy views are generated, and a recommender system will suffer a decline in performance. Thus, noisy views for recommendations are necessary to indicate the motivation of our work.

\section{Preliminaries}

\subsection{Problem Definition}
In collaborative filtering using Graph Neural Networks (GNNs), the user-item interactions are structured into a bipartite graph, denoted as $\mathcal{G}(\mathcal{U}, \mathcal{V}, \mathcal{E})$. In this model, $\mathcal{U} = \{u_1, u_2, ..., u_m\}$ represents the set of users, and $\mathcal{V} = \{v_1, v_2, ..., v_n\}$ stands for the set of items, with $m$ and $n$ indicating the respective counts of users and items. The edges $\mathcal{E}$ in the graph represent interactions between users and items. The primary objective of these recommender systems is to leverage a graph neural network to effectively capture and learn latent representations for both users and items. These representations are then utilized through the inner product to predict user preferences for upcoming items accurately.

\subsection{Noise-tolerant Symmetric Loss}
This section outlines the theoretical underpinnings of noise-tolerant symmetric loss for multi-class classification. In this context, let $\mathcal{X}$ represent the input space, and $\mathcal{Y} = \{1, \dots, k\}$ denote the set of class labels. Consider the unobserved clean training dataset $\mathcal{S} = \{x_i, y_{x_i}\}_{i=1}^N$, which consists of samples drawn independently and identically from an unknown distribution $\mathcal{D}$, with $N$ being the total number of samples. For a classifier function $f: \mathcal{X} \rightarrow \mathbb{R}$, let $\mathbb{E}_{x, y_x}(\mathbb{E}_{x, \hat{y}_x})$ and $\mathbb{E}_{\mathcal{D}}(\mathbb{E}_{\mathcal{D}_\eta})$ be considered equivalent. Under these conditions, the L-risk of classifier $f$ in a noise-free environment is defined as follows:
\begin{equation}
    R_L(f)=\mathbb{E}_{x,y_{x}}[(\mathcal{L}(f(x),y_x))]=\mathbb{E}_{\mathcal{D}}[(\mathcal{L}(f(x),y_x))],
    \label{eq10}
\end{equation}
where $\mathbb{E}$ represents the expectation, and $\mathcal{L}: \mathbb{R} \times \mathcal{Y} \rightarrow \mathbb{R}$ signifies the loss function. The noisy training set is denoted by $\hat{\mathcal{S}}=\{x_i, \hat{y}_{x_i}\}_{i=1}^N$, where the noisy labels $\hat{y}_{x_i}$ are defined as follows:
\begin{equation}
\begin{split}
     \hat{y}_{x_i}=\begin{cases}y_{x_i},\quad \mathrm{with\;probability}\; (1-\eta_{x_i})&
     \\
     i, \;i\in \{1,\ldots,k\},\;i \neq y_{x_i}\quad \mathrm{with\;probability}\;\bar{\eta}_{x_ii}&
     \end{cases},
\end{split}
\end{equation}
where $\eta_{x_i}$ and $\bar{\eta}_{x_ii}$ are the noisy ratios. When the noise exists, and the excepted risk under noisy data is:
\begin{equation}
R^{\eta}_\mathcal{L}(f)=\mathbb{E}_{\mathcal{D}_{\eta}}[ \mathcal{L}(f(x),\hat{y}_x)].
\label{eq12}
\end{equation}
{Existing work \cite{ghosh2017robust} demonstrates that if a loss function satisfies the symmetry condition, then it is noise-tolerant}. Specifically, a loss function $\mathcal{L}$ exhibits symmetry if it meets the following condition:
\begin{equation}
    \sum_{i=1}^k \mathcal{L}(f(x_i),y_{x_i})=c, \forall{x_i}\in\mathcal{X}, \forall{f},
    \label{eq13}
\end{equation}
where $c$ is a constant. The complete proof of the noise-tolerant symmetric loss is presented as follow.

\begin{proof}
For any $f$, we have,
\begin{equation}
\begin{split}
R^{\eta}_\mathcal{L}(f) & =\mathbb{E}_{x,\hat{y}}\mathcal{L}(f(x),\hat{y}) \\
& =\mathbb{E}_x\mathbb{E}_{y|x}\mathbb{E}_{\hat{y}|x,y} \mathcal{L}(f(x),\hat{y}) \\
& =\mathbb{E}_{\mathcal{D}}
\left[(1-\eta)\mathcal{L}(f(x),y) + \frac{\eta}{k-1}\sum_{i\neq y} \mathcal{L}(f(x_i),y_i)\right] \\
& = (1-\eta)R_\mathcal{L}(f) + \frac{\eta}{k-1}(c-R_\mathcal{L}(f)) \\
& =\frac{c\eta}{k-1}+\left(1-\frac{\eta k}{k-1}\right)R_\mathcal{L}(f)).
\end{split} 
\label{eq18}
\end{equation}
Let $C=\frac{c\eta}{(k-1)}$ be the constant item and $\alpha=(1- \frac{\eta k}{k-1})$ the coefficient related to the noisy probability $\eta$ and category number. Equation (\ref{eq18}) can be finally expressed as:
\begin{equation}
    R^{\eta}_\mathcal{L}(f)=C+\alpha R_\mathcal{L}(f).
\label{eq20}
\end{equation}
 Equation (\ref{eq20}) shows that $R^{\eta}_\mathcal{L}(f)$ is positively related to $R_\mathcal{L}(f)$ when the coefficient $\alpha>0$ (i.e. the noise ratio $\eta<\frac{k-1}{k}$). For any $f$, we hava,
\begin{equation}
R^{\eta}_{\mathcal{L}}(f^*)-R^{\eta}_{\mathcal{L}}(f) = \alpha(R_{\mathcal{L}}(f^*)-R_{\mathcal{L}}(f))\le0,
\label{eq9}
\end{equation}
where $f^*$ is a minimizer of $R_{\mathcal{L}}$. Equation (\ref{eq9}) demonstrates that the global minimizer $f^*$ of the $R^{\eta}_{\mathcal{L}}$ is same as the global minimizer $R_{\mathcal{L}}$. They show that if the noisy probability $\eta$ satisfies the condition that $\eta < \frac{k-1}{k}$ and the loss function is symmetric. The minimizer of the risk under noisy conditions approximates the minimization of the risk under clean conditions.
\end{proof}

\section{Methodology}
In Section 3, we explain the principles and the specific implementation of the SGCL method. It consists of five parts: (1) Symmetric Form of Contrastive Learning, (2) Symmetric Contrastive Loss, (3) Multi-task Training, (4) Complexity Analysis, and (5) Discussion.

\subsection{Symmetric Form of Contrastive Learning }
In this section, we build a connection between the noise-tolerant symmetric loss and the contrastive leaning (CL) objective. Then, we transfer CL into the symmetric form.
Inspired by the noise-tolerant symmetric loss framework \cite{ghosh2017robust}, graph contrastive learning can achieve excellent robustness against noisy views if it can satisfy the symmetry condition. In graph contrastive learning for collaborative filtering, given two generated views $z'$ and $z''$ of the same node, we define that the positive pair $(z',z'')$ with label $-1$ if they are the same distribution $(z',z'')\sim P_z$, and $1$ if the $(z',z'')$ do not follow the same distribution (e.g., in Figure \ref{fig2} (a) and Figure \ref{fig2} (b), the embedding distribution of $u'_1$ and $u''_1$ are different) $(z',z'')\sim P_{z'}P_{z''}$. 
This type of noisy view truly exists during the model training process. In existing works, such noisy views are automatically considered a positive pair (i.e., with the label -1) and directly ignored, which generates a suboptimal representation and decreases the recommendation accuracy. 

We employ the widely used contrastive loss, InfoNCE, to strengthen the consistency between positive pairs while diminishing it among negative pairs. This approach aids in improving representation learning, as illustrated below:
\begin{equation}
    \mathcal{L}_{ssl} = -\mathrm{log}\frac{e^{\phi(z',z'')/\tau}}{e^{\phi(z',z'')/\tau}+\sum_{j \in K}e^{\phi(z',z''_j)/\tau}},
    \label{eq7}
\end{equation}
where $\tau$ is the temperature coefficient, which is always greater than 0 and regulates the scaling. $z'$ and $z''$ represent distinct view representations of the same node derived from graph augmentation, while $z''_j$ denotes the negative sample for each node. The function $\phi(\cdot)$ calculates their similarity. The parameter $K$ corresponds to a batch of sampled nodes.
For ease of derivation, we define $s^+ = \phi(z', z'')/\tau$ and $s^- = \phi(z', z''_j)/\tau$. Consequently, Equation (\ref{eq7}) can be reformulated as follows:
\begin{equation}
 \mathcal{L}_{ssl}  =-\mathrm{log}\frac{e^{s^+}}{e^{s^+} + \sum_{j\in K}e^{s-}}, \\
\end{equation}
where $s^+$ and $s^-$ denote the score of the positive pair and the score of the negative pair, respectively. The InfoNCE loss can be seen as a variant of the ($K + 1$)-way softmax cross-entropy loss, essentially aiming to classify whether a pair $(z',z'')$ is a negative or positive sample by maximizing the positive score $s^+$ and minimizing the negative score $s^-$. Therefore, when considering noisy views, InfoNCE can be viewed as a binary classification task where the labels are influenced by noise. In this context, Equation (\ref{eq13}) can be rephrased to articulate the symmetry condition as follows:
\begin{equation}
    \mathcal{L}\left(f\left(x\right),y_x\right) + \mathcal{L}\left(f\left(x\right),\hat{y}_x\right) = c,
    \label{eq:10}
\end{equation}
where $y_x$ and $\hat{y}_x$ denote the true label and noisy label, typically taking values of 1 and -1 respectively, where $\hat{y}_x=y_x$ with the probability $1-\eta$ and $\hat{y}_x=-y_x$ with the noisy probability $\eta$. 
It is worth mentioning that the symmetry condition should also be satisfied in the gradients.
Hence, the contrastive learning objective is symmetric when there exists a loss function $\mathcal{L}$ satisfying the symmetry condition, adhering to the following structure: 
\begin{equation}
\mathcal{L} = \underbrace{\mathcal{L}(s^+,y)}_{Positive\;pair} + \; \lambda \underbrace{\sum^K\mathcal{L}(s^-,\hat{y})}_{K\;Negative\;pairs},
\end{equation}
where $\lambda>0$ is the weight term. Nevertheless, InfoNCE is not a symmetric loss function, as it fails to meet the symmetry condition in the gradient. The detailed derivation is illustrated below:
\begin{equation}
\begin{split}
    & \frac{\partial\mathcal{L}_{ssl}}{\partial s^+}=\frac{-1}{\mathcal{L}_{ssl}}\cdot\frac{e^{s+} \cdot \sum^{K}_{j=1}e^{s^-}}{(e^{s+} + \sum^{K}_{j=1}e^{s^-})^2}, \\
    &  \frac{\partial\mathcal{L}_{ssl}}{\partial s^-}=\frac{-1}{\mathcal{L}_{ssl}}\cdot\frac{e^{s+}(1-e^{s^-}) + \sum^{K}_{j=1}e^{s^-}}{(e^{s+} + \sum^{K}_{j=1}e^{s^-})^2}.
\end{split}
\end{equation}
We compute the gradients of $s^+$ and $s^-$ in a batch of data where the sum is not constant, leading to a failure to satisfy the symmetry condition. The limitations of InfoNCE inspire us to explore a new contrastive loss that satisfies the symmetry condition.

\subsection{Symmetric Contrastive Loss}
This work directly proposes the symmetric contrastive loss (SCL), which is a variant of InfoNCE \cite{chuang2022robust}. The SCL is shown as follows:
\begin{equation}
\mathcal{L}_{SCL}=-\frac{e^{ps^+}}{p}+\frac{\lambda (-e^{s^+}+\sum^{K}_{j=1}e^{s^-})^p}{p},
\end{equation}
where $p$ and $\lambda$ (both in the range of $(0,1]$) are utilized to find an equilibrium between convergence speed and robustness. A significant point to note is that SCL does not necessitate a specific noisy estimator in an explicit form. Instead, it leverages the scores $s^+$ and $s^-$, along with their relationship (which the loss function quantifies), to function as noise estimations. When $p=1$, the SCL satisfies the symmetry property and achieves robustness against noisy contrastive views with \textbf{$\mathcal{L}(y,s)=ye^s$}, which is formulated as:
\begin{equation}
\mathcal{L}_{SCL}=-(1+\lambda)e^{s^+}+\sum^K_{j=1}e^{s^-}.
\end{equation}
We provide theoretical proof of robustness against the noisy view of SCL with $p \rightarrow 1$ and the exponential loss.
\begin{Corollary}
Let $f^*_{\eta}=\mathrm{arginf}_{f \in \mathcal{F}}R^{\eta}_{\mathcal{L}}(f)$ be the minimizer of noisy risk, $f=\mathrm{arginf}_{f \in \mathcal{F}}R_{\mathcal{L}}(f)$ be the minimizer of optimal risk, and $\delta$ be $\mathbb{E}_{\mathcal{D}}[\mathcal{L}(f(x),y_x)]$, if the $\eta \le \eta_{max} < 0.5$, we have $R^{\eta}_{\mathcal{L}}(f^*_{\eta}) \le \frac{\delta}{1-2\eta_{max}} \nonumber.$
\end{Corollary}
\begin{proof}
Giving the exponential loss $\mathcal{L}(f(x),y_x)=ye^{f(x)}$ that satisfies the symmetry property,
\begin{equation}
    \mathcal{L}(f(x),y_x)+ \mathcal{L}(f(x),\hat{y}_x)=0.\nonumber
\end{equation}
Consider a binary classification, the excepted risk under noisy data,
\begin{equation}
\begin{split}
    R^{\eta}_\mathcal{L}(f) & =\mathbb{E}_{x,\hat{y_x}}\left[\mathcal{L}\left(f(x),\hat{y_x}\right)\right] \\
& =\mathbb{E}_{\mathcal{D}}\left[(1-\eta)\mathcal{L}(f(x),y_x) + \eta(-\mathcal{L}(f(x),y_x)) \right] \\
& =\mathbb{E}_{\mathcal{D}}[(1-2\eta) \mathcal{L}(f(x),y_x)] \\
& = (1-2\eta)R_L(f),
\end{split}
\end{equation}
since $f=\mathrm{arginf}_{f \in \mathcal{F}}R_{\mathcal{L}}(f)$ and $f^*_{\eta}=\mathrm{arginf}_{f \in \mathcal{F}}R^{\eta}_{\mathcal{L}}(f) $, we have
\begin{equation}
   R^{\eta}_\mathcal{L}(f^*_{\eta}) - R^{\eta}_\mathcal{L}(f) =(1-2\eta)(R_\mathcal{L}(f^*_{\eta})-R_\mathcal{L}(f)) \le 0 ,
\end{equation}
according to Equation (\ref{eq10}) and  (\ref{eq12}), we have
\begin{equation}
     \mathbb{E}_{\mathcal{D}}[(1-2\eta)(\mathcal{L}(f^*_{\eta}(x),y_x)-\mathcal{L}(f(x),y_x))] \le 0 ,
\end{equation}
since $f^*_{\eta}$ is the minimizer of $R^{\eta}_\mathcal{L}$, this suggests that
\begin{equation}
\mathbb{E}_{\mathcal{D}}[(1-2\eta)(\mathcal{L}(f^*_{\eta}(x),y_x))] \le \mathbb{E}_{\mathcal{D}}[(1-2\eta)(\mathcal{L}(f(x),y_x)) \nonumber
\end{equation}
\begin{equation}
    \Longrightarrow \underset{\eta}{\mathrm{min}}(1-2\eta)\mathbb{E}_{\mathcal{D}}[\mathcal{L}(f^*_{\eta}(x),y_x)]\le \mathbb{E}_{\mathcal{D}}[\mathcal{L}(f(x),y_x)].
\end{equation}
Since $\delta = \mathbb{E}_{\mathcal{D}}[\mathcal{L}(f(x),y_x)]$ and the loss is non-negative, we have 
\begin{equation}
    \underset{\eta}{\mathrm{min}}(1-2\eta)\mathbb{E}_{\mathcal{D}}[\mathcal{L}(f^*_{\eta}(x),y_x)] \le \delta 
\end{equation}
\begin{equation}
    \Longrightarrow R^{\eta}_{\mathcal{L}}(f^*_{\eta}) \le \frac{\delta}{1-2\eta_{max}},
\end{equation}
where $\eta_{max} = \mathrm{sup}\;\eta$ and $\eta \le \eta_{max} < 0.5$. This completes the proof.
The above proof implies that when $p \rightarrow 1$, the SCL achieves robustness against noisy views.
\end{proof}

It is commonly acknowledged that InfoNCE serves as a variational lower bound for expressing mutual information (MI) using KL-divergence. Nevertheless, previous studies have illustrated theoretical constraints in estimating MI through KL-divergence \cite{chuang2022robust}. It reacts to slight fluctuations in data instances but disregards the geometric attributes of the intrinsic data distributions. Consequently, the encoder might struggle to acquire comprehensive and precise representations, as even minor discrepancies are capable of maximizing the KL divergence. The Wasserstein Dependency Measure (WDM), based on the Wasserstein distance \cite{panaretos2019statistical}, offers an advantageous alternative to KL divergence for estimating mutual information. WDM is presented as follows:
\begin{equation}
    \mathcal{W}(p,q) = \mathop{\mathrm{inf}}_{\gamma \in \Phi(p,q)} \mathbb{E}_{(x,y)  \sim \gamma, (x',y') \sim \gamma}\left [ \Vert x-y\Vert + \Vert x' -y'\Vert\right ], 
\end{equation}
where $\Phi(p,q)$ denotes the set of all possible joint distributions combining distributions $p$ and $q$.
We demonstrate that the SCL functions as a lower bound in contrast to mutual information (MI) articulated through the WDM, which is displayed as:
\begin{equation}
\begin{split}
       -{\mathbb{E}[\mathcal{L}^{\lambda, p=1}_{SCL}]} &\le C \cdot \mathcal{I}_{\mathcal{W}}(z';z'') \\
       &:= C \cdot \mathcal{W}(P(z',z''),P(z')P(z'')),
\end{split}
\end{equation}
where $C$ is a constant and $z'$ and $z''$ are the learned representations. The complete proof process is shown below:

\begin{proof}
    We can bound the symmetric loss using the additivity of expectation as follows \\
    \begin{equation}
    \begin{aligned}
        -\mathbb{E}[\mathcal{L}^{\lambda, p=1}_{SCL}(s)]
        & =\mathbb{E}_{(z', z'') \sim P_{z'z''}}\left[(1+\lambda)e^{\phi(z',z'')/\tau}\right] - \mathbb{E}_{z' \sim P_{z'}, z'' \sim P_{z''}}\left[\lambda\sum^{K}_{j=1}e^{\phi(z',z''_j)/\tau}\right] \\ 
        &=\mathbb{E}_{(z', z'') \sim P_{z'z''}}\left[(1+\lambda)e^{\phi(z',z'')/\tau}\right] - \lambda K \mathbb{E}_{z' \sim P_{z'}, z'' \sim P_{z''}}\left[e^{\phi(z',z'')/\tau}\right]  \\
        &\le (1+\lambda)(\mathbb{E}_{(z', z'') \sim P_{z'z''}})\left[ e^{\phi(z',z'')/\tau}\right] -
        \mathbb{E}_{z' \sim P_{z'}, z'' \sim P_{z''}}\left[e^{\phi(z',z'')/\tau}\right],
        \end{aligned}
    \end{equation}
since $s=\phi(z',z'')$ represents the similarity score between two embedding vectors, we have $-\frac{1}{\tau} \le s \le \frac{1}{\tau}$ and $|\nabla_s e^s| \le e^{1/\tau}$.
We expand $e^{\phi(\cdot)}$ using the inequality form of the mean value theorem. We have
\begin{equation}
|e^{\phi(f(x_1),f(x_2))/\tau}-e^{\phi(f(x'_1),f(x'_2))/\tau}| 
\le \frac{e^{1/\tau}}{\tau}|\left<\phi(f(x_1)), \phi(f(x_2))\right> - \left<\phi(f(x'_1)),\phi(f(x'_2))\right>|,
\label{eq:24}
\end{equation}
where $\left< \cdot \right>$ denotes the vector inner product. $x_1$, $x_2$, $x'_1$, and $x'_2$ are sampled from the same distribution.
\begin{equation}
\begin{aligned}
   &|\left<\phi(f(x_1)), \phi(f(x_2))\right> - \left<\phi(f(x'_1)),\phi(f(x'_2))\right>| 
   \\ 
   &= |\left<\phi(f(x_1)) - \phi(f(x'_1)), \phi(f(x_2))\right> + \left<\phi(f(x'_1)), \phi(f(x_2)) - \phi(f(x'_2))\right>|
   \\
   &\le |\left<\phi(f(x_1)) - \phi(f(x'_1)), \phi(f(x_2))\right>| + |\left<\phi(f(x'_1)), \phi(f(x_2)) - \phi(f(x'_2))\right>| \\
    &\le \Vert \phi(f(x_1)) - \phi(f(x'_1))\Vert \Vert\phi(f(x_2))\Vert+ \Vert\phi(f(x'_1)) \Vert
    \Vert \phi(f(x_2)) - \phi(f(x'_2)) \Vert \\
    &=\Vert \phi(f(x_1)) - \phi(f(x'_1))\Vert  + 
    \Vert \phi(f(x_2)) - \phi(f(x'_2)) \Vert \\
    & =Lip(\phi) (\Vert f(x_1) - f(x'_1)\Vert  + 
    \Vert f(x_2) - f(x'_2) \Vert)
    \\
    &=Lip(\phi) d((f(x_1),f(x'_1)),
    (f(x_2),f(x'_2)).
   \end{aligned}
   \label{eq:25}
\end{equation}
Equations (\ref{eq:24}) and (\ref{eq:25}) clearly show that the Lipschitz constant of $e^{\phi(\cdot)}$, with respect to the metric $d$, is constrained by the constant $\frac{Lip(\phi) \cdot e^{1/\tau}}{\tau}$. Additionally, we have
\begin{equation}
\begin{aligned}
    -\mathbb{E}[\mathcal{L}^{\lambda, p=1}_{SCL}(s)]  
    \le (1+\lambda)(\mathbb{E}_{(z', z'') \sim P_{z'z''}})\left[ e^{\phi(z',z'')/\tau}\right] -
        \mathbb{E}_{z' \sim P_{z'}, z'' \sim P_{z''}}\left[e^{\phi(z',z'')/\tau}\right]
    \end{aligned}
    \nonumber
\end{equation}
\begin{equation}
       \le C \cdot \mathcal{W}(P_{z',z''}, P_{z'}P_{z''}),
\end{equation}
where $C = \frac{(1+\lambda)Lip(f)e^{1/e}}{\tau}$ is a constant item. 
\end{proof}
SGCL provides a dependable perspective due to its resilience to noisy views. In contrast to KL divergence, WDM evaluates the density ratio between two distributions and quantifies the cost of transitioning from one distribution to another. This reliability of SGCL allows it to calculate the divergence between two distributions without exhibiting excessive sensitivity to noisy views at the individual sample level, given that such noise does not fundamentally modify the distributions' geometrical properties.

\begin{algorithm}[t]
\caption{The algorithmic workflow of SGCL}
\begin{algorithmic}[1]
\Require User-item interaction graph $\mathcal{G}(\mathcal{U}, \mathcal{V}, \mathcal{E})$, hyperparameter $\mathcal{\lambda}$, $\mathcal{\beta}$, and $p$
\Ensure parameters $\Theta$ of SGCL

\State Initialize the parameters $\Theta$ of SGCL randomly
\For{each training epoch}
    \State \texttt{// Forward propagation process}
    \State Generate two perturbed graphs $\mathcal{G}_1$ and $\mathcal{G}_2$ from origin graph $\mathcal{G}$
    \State $e_u$, $e_i \gets$ GNN
    ($\mathcal{G}$, $e^{(0)}_u$, $e^{(0)}_i$)
    \State  $z'_u$, $z'_i$ $\gets$
    GNN ($\mathcal{G}_1$, $z^{(0)}_u$, $z^{(0)}_i$)
    \State $z''_u$, $z''_i$ $\gets$
    GNN ($\mathcal{G}_2$, $z^{(0)}_u$, $z^{(0)}_i$)
    \State \texttt{// Calculate Loss}
    \State Compute the contrastive learning $\mathcal{L}_{SCL}$ according to Equation (\ref{eq17})
    \State Compute the main recommendation loss $\mathcal{L}_{main}$  according to Equation (\ref{eq14})
    \State \texttt{// Backward propagation process}
    \State Update the GNN parameters $\Theta$
\EndFor
\State \Return model parameters $\Theta$
\end{algorithmic}
\label{al: 01}
\end{algorithm}

\subsection{Multi-task Training}
We implement SGCL based on the backbone SGL (using the edge dropout augmentation operator). Specifically, we utilize symmetric contrastive loss instead of InfoNCE in SGL to facilitate the learning of user and item embeddings. We also adopt a multi-task training~\cite{zhang2021survey,adatask} strategy to optimize the loss function concurrently:
\begin{equation}
    \mathcal{L} = \mathcal{L}_{main} + \beta\mathcal{L}_{SCL},
    \label{eq14}
\end{equation}
where $\beta$ acts as a hyperparameter that controls the scale of $\mathcal{L}_{SCL}$. The BPR loss, the primary recommendation loss, is used to model the user's preference between two items. The BPR loss is expressed as follows:
\begin{equation}
    \mathcal{L}_{main}=-\sum_{(u,i,j)\in\mathcal{O}}{\log}\sigma \left ( y_{ui} - y_{uj} \right) + \alpha\Vert \Theta \Vert_2^2,
    \label{14}
\end{equation}
where $\mathcal{O}=\{(u,i,j) | (u,i) \in \mathcal{O}^+, (u,j) \in \mathcal{O}^-\}$ represents the training dataset, $\mathcal{O}^+$ signifies the observed interactions, and $\mathcal{O}^-$ = $(\mathcal{U} \times \mathcal{V}) / \mathcal{O}^+$ indicates the unobserved interactions. The symbol $\Theta$ denotes the set of model parameters, and $\alpha$ refers to the hyperparameter. 
The contrastive objective often serves as a secondary task to enhance the learning of representations. The symmetric contrastive loss for user nodes, $\mathcal{L}^{user}_{SCL}$, is described as follows:
\begin{equation}
    \mathcal{L}^{user}_{SCL} = -\frac{e^{s^+_{u}p}}{p} + \frac{\lambda(-e^{s^+_{u}}+\sum^k_{j=1}e^{s^-_{u}})^p}{p},
    \label{eq16}
\end{equation}
where $s^+_{u}=\phi(z'_{u},z''_{u})/\tau$ and $s^-_{u}=\phi(z'_u,z''_j)/\tau$ represent the positive and negative sample scores for a user node, respectively, similarly, the contrastive objective for item nodes, $\mathcal{L}^{item}_{SCL}$, is calculated comparably. We amalgamate these two losses to determine the composite self-supervised task as follows:
\begin{equation}
    \mathcal{L}_{SCL} = \mathcal{L}^{user}_{SCL} + \mathcal{L}^{item}_{SCL}.
    \label{eq17}
\end{equation}
We provide a formal description of the training procedure in the Algorithm \ref{al: 01}
.
\subsection{Complexity Analysis}
In this section, we thoroughly analyze the time complexity of the SGCL to assess the model's efficiency. Given that SGL serves as the underlying framework and we only alter the loss function, the time complexity aligns with that of SGL. Specifically, the requirement for graph augmentations incurs a time complexity of $\mathcal{O}(2\times|\mathcal{E}|+4\sigma\times|\mathcal{E}|)$. Furthermore, the graph convolution stage demands $\mathcal{O}((2+4\sigma)\times l \times d \times|\mathcal{E}|)$ time. The BPR loss and SSL loss require $\mathcal{O}(2 \times B \times d)$ and $\mathcal{O}(B\times d + B \times K \times d)$ time, respectively. Here, $\mathcal{E}$ signifies the count of edges in the bipartite graph, $d$ indicates the embedding dimension, $\sigma$ denotes the edge dropout rate during data augmentation, $B$ is the batch size, $l$ is the number of layers, and $K$ represents the number of nodes per batch.  

\subsection{Discussion}
Recently, RINCE \cite{chuang2022robust} employs the symmetric property to design robust contrastive learning loss to improve the model's robustness against various noise types in images and videos. However, this work has two key distinctions (i.e., contributions) between this work and RINCE. \textit{On the one hand, we focus on the noise generated when manually augmenting the user-item graph in recommendation systems}. In detail, existing works \cite{yu2022graph, xia2023automated, 10158930} point out that dropping the key nodes
and the associated edge will distort the original graph. We design experiments to investigate how removing key nodes and edges affects recommendation performance. We experimentally validate that removing key edges or nodes generates noisy views defined in Section \ref{section 2.1}, and ignoring these noisy views directly will impair recommendation performance.
\textit{On the other hand, although we all adopt symmetry theory to design new loss functions, our proposed loss function is different from RINCE}. Specifically, our designed noise-tolerant contrastive loss fully satisfies symmetry with $\mathcal{L}(f(x),y_x)=ye^{f(x)}$. We provide a theory demonstration that SCL exhibits strong robustness against noisy views. As a connection, we follow the proof technique from RINCE to show that SCL serves as a lower bound of WDM and provides robustness against noisy views.

\begin{table*}[htpb]
    \centering
    \caption{The comparative experiments are conducted on the Yelp2020 dataset. In the table, the best baseline is indicated with an underscore, while the overall best result is highlighted in bold. 'Improv.' denotes the percentage improvement in performance compared to the best baseline.}
    \begin{tabular}{c|cccccc}
    \hline
    \hline
          \rowcolor{gray!20} Datasets &  \multicolumn{6}{c}{Yelp2020}  \\ \hline
          \rowcolor{gray!20}Methods &  Precision@10 & Recall@10 & NDCG@10 & Precision@20 & Recall@20 & NDCG@20 \\  \hline
          BasicMF & 0.0199  &0.0375 & 0.0325 & 0.0169 &0.0368 & 0.0418\\
          LightGCN & 0.0261& 0.0494 & 0.0436 & 0.0221 & 0.0818 & 0.0550  \\
          LightGCL &0.0276 & 0.0512 & 0.0447 & 0.0231 & 0.0853 & 0.0568 \\
          NCL & 0.0297 & 0.0557 & 0.0470 & 0.0247& 0.0912 & 0.0621\\
          DCCF & 0.0295 &0.0541 & 0.0478 & 0.0244 &0.0892 &0.0613 \\
          SimGCL  &0.0321 &0.0596 & 0.0534 &0.0269 &0.0969 & 0.0663\\
          SGL &0.0311 &0.0581 & 0.0516 &0.0256  &0.0947 & 0.0644\\
          HCCF &  0.0294 & 0.0557 & 0.0500 & 0.0242& 0.0900 &0.0620\\
          AdaGCL & 0.0323& 0.0590 & 0.0517 & \underline{0.0260} & 0.0948 & 0.0649 \\
          AutoCF & 0.0320 &0.0592 &0.0525 & 0.0258 &0.0936 & 0.0654\\
          GFormer & \underline{0.0328}  & \underline{0.0603} & \underline{0.0534} & \underline{0.0260} & \underline{0.0974} & \underline{0.0664}\\
          \hline
          \rowcolor{gray!20} SGCL & \textbf{0.0344} &  \textbf{0.0645} & \textbf{0.0574} &\textbf{0.0279} & \textbf{0.1033} & \textbf{0.0710}\\
          Improv. & 4.88\%& 6.97\% & 7.49\% & 2.95\% & 6.06\% & 6.93\%\\
          \hline
          \hline
    \end{tabular}
    \label{tab: 5.1}
\end{table*}

\begin{table*}[htpb]
    \centering
    \caption{The comparative experiments are conducted on the Amazon-CD dataset. In the table, the best baseline is indicated with an underscore, while the overall best result is highlighted in bold. 'Improv' denotes the percentage improvement in performance compared to the best baseline.}
    \begin{tabular}{c|cccccc}
    \hline
    \hline
          \rowcolor{gray!20} Datasets &  \multicolumn{6}{c}{Amazon-CD}  \\ \hline
         \rowcolor{gray!20} Methods &  Precision@10 & Recall@10 & NDCG@10 & Precision@20 & Recall@20 & NDCG@20 \\  \hline
          BasicMF & 0.0282 & 0.0724 & 0.0568 & 0.0220& 0.1152 & 0.0928  \\
          LightGCN & 0.0336 & 0.0900 & 0.0690 & 0.0258 &0.1368 & 0.0864 \\
          LightGCL &0.0351 &0.0976 & 0.0769 &0.0267 &0.1468 & 0.0928\\
          NCL &0.0357 & 0.0715 &0.0769 & 0.0275 & 0.1464 & 0.0922 \\
          DCCF &  0.0374 & 0.0999  & 0.0796 & 0.0283 & 0.1494 & 0.0954 \\
          SimGCL & 0.0384 & 0.1060 & 0.0847 & 0.0289 &0.1563 & 0.1003 \\
          SGL & 0.0398 & 0.1063 & 0.0768 & 0.0299 &0.1534 & 0.1016 \\
          HCCF & 0.0392 & 0.1029 & 0.0818 & 0.0290 &0.1502 & 0.0970\\
          AdaGCL & 0.0399& 0.1086 & 0.0821 & 0.0294 & 0.1544 & 0.0100 \\
          AutoCF & \underline{0.0408} & 0.1087 & \underline{0.0883} & \underline{0.0302} &0.1571 & \underline{0.1038}\\
          GFormer & 0.0400 & \underline{0.1088} & 0.0877 & 0.0297 & \underline{0.1583} &0.1036\\
          \hline
          \rowcolor{gray!20}SGCL & \textbf{0.0445} & \textbf{0.1186} & \textbf{0.0952} & \textbf{0.0339} & \textbf{0.1741} & \textbf{0.1132}\\
          Improv. &  9.07\% &9.01\% &7.81\% & 12.25\% & 9.98\% & 9.06\%\\
          \hline
          \hline
    \end{tabular}
    \label{tab: 5.2}
\end{table*}
\begin{table*}[htpb]
    \centering
    \caption{The comparative experiments are conducted on the Amazon-Book dataset. In the table, the best baseline is indicated with an underscore, while the overall best result is highlighted in bold. 'Improv' denotes the percentage improvement in performance compared to the best baseline.}
    \begin{tabular}{c|cccccc}
    \hline
    \hline
          \rowcolor{gray!20} Datasets &  \multicolumn{6}{c}{Amazon-Book}  \\ \hline
          \rowcolor{gray!20} Methods &  Precision@10 & Recall@10 & NDCG@10 & Precision@20 & Recall@20 & NDCG@20 \\  \hline
          BasicMF & 0.0385 &0.0592 & 0.0505 & 0.0311& 0.0505 &0.0709 \\
          LightGCN & 0.0503 &0.0820 & 0.0945 & 0.0392 & 0.1234 & 0.0953\\
          LightGCL &0.0482 &0.0743 &0.0733 & 0.0385& 0.1170 & 0.0884 \\
          NCL & 0.0566 & 0.0884 & 0.0885 & 0.0440 & 0.1388 & 0.1041 \\
          DCCF & 0.0446 & 0.0939 & 0.0745 & 0.0359& 0.1050 &0.0796  \\
          SimGCL & 0.0501 & 0.0935 & 0.0904 & 0.0453 &0.1411 & 0.1104 \\
          SGL & 0.0512 & 0.0904 & 0.0810 & 0.0425 &0.1384 & 0.1068\\
          HCCF & 0.0522 & 0.0805 & 0.0958 & 0.0411 & 0.1263 & 0.1120 \\
          AdaGCL & 0.0600 & 0.0912 & 0.0934  & 0.0455 &0.1397 & 0.0110\\
          AutoCF & \underline{0.0614} & \underline{0.0950} & \underline{0.0970} & \underline{0.0469} & \underline{0.1412} & \underline{0.1123} \\
          GFormer & 0.0603 & 0.0899 & 0.0916 & 0.0450& 0.1340 & 0.1063  \\
          \hline
          \rowcolor{gray!20} SGCL & \textbf{0.0675} & \textbf{0.1050} & \textbf{0.1054} & \textbf{0.0520} & \textbf{0.1578} & \textbf{0.1232}\\
          Improv. & 9.93\% & 10.53\% & 8.66\% &10.87\% & 11.76\% & 9.71\% \\
          \hline
          \hline
    \end{tabular}
    \label{tab: 5.3}
\end{table*}

\section{Experiments}
In this section, we conduct a series of experiments to verify SGCL and address the key research questions outlined below:
\begin{itemize}[leftmargin=*]
\item\textbf{RQ1}: How does the performance of our SGCL  compare to other state-of-the-art (SOTA) recommendation baselines?
\item\textbf{RQ2}: Does SGCL demonstrate increased robustness in handling noisy and limited data compared to baseline methods?
\item \textbf{RQ3}: How influential are the primary elements of SCL within our model?
\item\textbf{RQ4}: What is the impact of adjusting the key hyperparameters on the performance of SGCL?
\end{itemize}
\subsection{Experiment settings}
\subsubsection{Evaluation Datasets and Metrics} 
We assess our SGCL model using three real-world datasets: yelp2020, amazon-cd, and amazon-book\footnote{https://nijianmo.github.io/amazon/index.html}. Details of these datasets are summarized in Table~\ref{tab:datasets}. 
To measure the accuracy of our recommendations, we use two comprehensive ranking evaluation protocols. Specifically, we calculate Recall@$N$ and Normalized Discounted Cumulative Gain (NDCG)@$N$ to evaluate recommendation accuracy, with cutoffs established at 10 and 20.
\subsubsection{Comparison with Baselines}
We assess the effectiveness of SGCL by comparing it with the following baseline methods:
\begin{itemize}[leftmargin=*]
    \item \textbf{BasicMF} \cite{wang2012nonnegative}: A traditional matrix factorization method maps users and items into latent vectors.
    \item \textbf{LightGCN} \cite{he2020lightgcn}: It is a potent collaborative filtering method based on graph convolutional networks (GCNs) that simplifies NGCF's message propagation scheme by removing non-linear projection and activation.
    \item \textbf{NCL} \cite{lin2022improving}: 
    It integrates structural and semantic neighbors into contrastive pairs for graph collaborative filtering, formulating structure and prototype-contrastive objectives for implementation.
    \item \textbf{DCCF} \cite{ren2023disentangled}: This method employs a disentangled contrastive collaborative filtering approach for recommendations. It facilitates adaptive interaction augmentation by leveraging disentangled user (item) intent-aware dependencies that are learned dynamically.
    \item \textbf{SimGCL} \cite{yu2022graph}: This model implements a straightforward contrastive learning (CL) method that eschews graph augmentations in favor of introducing uniform noise into the embedding space to create contrastive views.
    \item \textbf{HCCF} \cite{xia2022hypergraph}: This architecture employs a hypergraph-enhanced, cross-view contrastive learning framework to simultaneously capture both local and global collaborative relationships.
    \item \textbf{LightGCL} \cite{caisimple}: This is an effective graph contrastive that uses the singular value decomposition for contrastive augmentation.
    \item \textbf{AdaGCL} \cite{Jiang2023adagcl}
    : It uses graph generative and denoising models to enhance recommendation systems by learning robust node embeddings from adaptive views of large-scale graphs.
    \item \textbf{GFormer} \cite{Li2023gformer}: It utilizes rationale-aware generative semi-supervised learning to automate the self-supervised augmentation process, effectively distilling significant patterns of user-item interactions.
    \item \textbf{AutoCF} \cite{xia2023automated}: This model adapts well to data augmentation and demonstrates robustness to noise. It features a graph augmentation strategy that automatically distills valuable self-supervision information through perturbations.
\end{itemize}

\begin{table}[t]
\centering
 \caption{Detailed statistics for each dataset.\label{tab:datasets}}
 \renewcommand{\arraystretch}{1} 
    \begin{tabular}{c|ccccl}
     \hline
     \hline
    \rowcolor{gray!20} Dataset & \#Users  & \#Items & \#Interactions & Density \\ 
    \hline
    Yelp2020 & 71,135 &  45,063 & 1,430,480 & $4.6\times10^{-4}$\\
    Amazon-CD & 22,947& 18,394& 328,779& $7.9\times10^{-4}$  \\
    Amazon-Book & 52,406 & 41,263 & 1,464,454 & $6.7\times 10^{-4}$ \\
    \hline
    \end{tabular}
    \label{tab01_}
\end{table}
\subsubsection{Hyperparameter Settings}
We developed the SGCL using the SELFRec recommendation library in PyTorch \cite{yu2023self}. For parameter optimization, we use the Adam optimizer with default settings including a learning rate of 0.001, a batch size of 4,096, and an embedding size of 64. We select the number of propagation layers from the set $\{1,2,3\}$. The penalty coefficient $\beta$ in Equation \ref{eq14} is determined from the range $\{1,1e^{-1},1e^{-2},1e^{-3},1e^{-4},1e^{-5}\}$. Both the penalty coefficient $\lambda$ and the hyperparameter $p$ in Equation \ref{eq16} are fixed at 0.01. Following the guidelines from the SGL paper, we set the temperature $\tau$ to 0.2. For baseline comparisons, we initially use the best hyperparameter configurations cited in the baseline studies and subsequently refine all hyperparameters through grid search.

\subsection{Overall Performance Comparison (RQ1)}
Tables \ref{tab: 5.1}, \ref{tab: 5.2}, and \ref{tab: 5.3} present a detailed comparison of the performance of all baseline models across three datasets. From these results, we can draw the following main observations:

Our proposed SGCL framework achieves notable performance enhancements compared to existing state-of-the-art recommendation baselines. Specifically, the improvement ranges from $2.95\%$ to $12.25\%$ in all metrics. The notable improvement can be attributed to the efficient and robust SGCL framework. By proposing the symmetric contrastive loss, which serves as a lower bound of mutual information expressed through WDM, SGCL proves to be robust against noisy contrastive views. From another perspective, this compels the model to acquire superior user and item representations, consequently yielding better recommendation results.  

Upon a more detailed analysis of the experimental results, it becomes apparent that self-supervised contrastive learning methods (e.g., SGL, SimGCL, and NCL) outperform supervised approaches  (e.g., BasicMF and LightGCN). This robust performance demonstrates the importance and effectiveness of self-supervised data augmentation auxiliary tasks in graph collaborative filtering. We analyze the underperformance of supervised methods and find that it may be attributed to the scarcity of labeled data in recommendation tasks. 
Self-supervised learning can mitigate this problem by generating extra supervisory signals from limited data interactions. For example, SGL employs random data augmentation directly on the interaction graph to generate self-supervised signals. SimGCL directly injects noise into embeddings to create contrasting views.

Although  SSL recommendation methods have improved performance (e.g., SGL), these adaptive contrastive learning methods outperform SSL (e.g., AutoCF and GFomer). Applying random augmentation directly to the user-item interaction graph may disrupt the overall graph connectivity, leading to less effective representations of user-item relationships. However, adaptive contrastive learning frameworks alleviate this issue by intelligently generating contrasting views through intricately designed network structures. The contrastive loss function introduced in this paper, which is tolerant to noise, successfully reduces the negative impact of noisy views on recommendation performance, further improving the accuracy of recommendations. The experimental results on three datasets indicate that SGCL exhibits some improvements compared to adaptive data augmentation methods, further confirming the effectiveness of SGCL in enhancing recommendation performance.

\begin{figure}[t]
\setlength{\belowcaptionskip}{-0.1cm}
	\centering
         
        \vspace{1pt}
\begin{minipage}
{0.9\linewidth}
\centerline{\includegraphics[width=\textwidth]{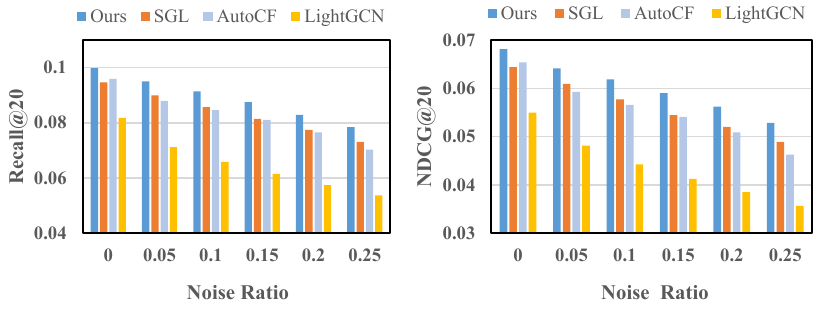}}
    \centerline{\;\;\;\;\;\;\;\;(a) Yelp2020}
\end{minipage}
\begin{minipage}
{0.9\linewidth}
\centerline{\includegraphics[width=\textwidth]{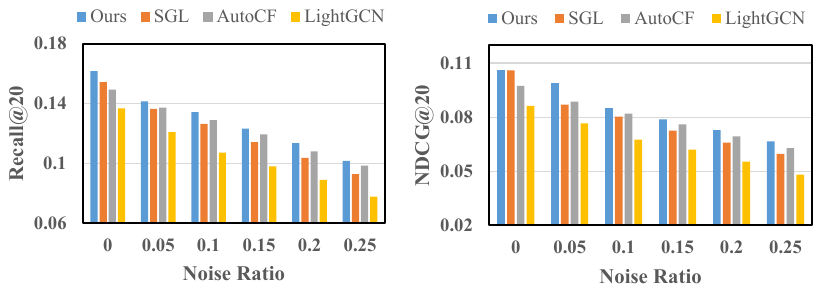}}
    \centerline{\;\;\;\;\;\;\;(b) Amazon-CD}
\end{minipage}
	\caption{Model performance comparison with different noise ratios. We sequentially replace genuine user-item interactions with 5\%, 10\%, 15\%, 20\%, and 25\% synthetic user-item edges.}
	\label{fig3}
\end{figure}

\subsection{Robustness Analysis (RQ2)}
\label{sc 5.2}
This section validates the model's robustness from two perspectives: noisy interactions and limited datasets. The experimental results indicate that our proposed SGCL demonstrates better robustness than other models.

For a comprehensive Evaluation Of the model's robustness to noisy interactions, we maintain the test dataset unchanged and replace genuine user-item interaction edges in the training dataset with fake interactions at varying proportions. Specifically, we generate corrupted user-item graphs by replacing 5\%, 10\%, 15\%, 20\%, and 25\% of the genuine user-item interaction edges. We compare the performance of our SGCL with other models, including SGL, LigthGCN, and AutoCF, on the Amazon-CD and Yelp2020 datasets. The results of these experiments are presented in Figure \ref{fig3}.

Figure \ref{fig3} shows that our SGCL outperforms the other baseline models on both datasets.
Our findings reveal that as the noise ratio increases, the performance of supervised learning methods deteriorates significantly. In contrast, all self-supervised methods, including SGCL, demonstrate superior performance at the same percentage of data noisy ratio. This underscores the stronger robustness of self-supervised methods compared to supervised methods within a specific noisy ratio range. Furthermore, our results show that SGCL outperforms the other two self-supervised methods, a performance advantage attributed to the SCL employed in SGCL, which aids the model in learning better representations with the noise label. These findings are further supported by the ablation results in Section 4.4. 

\subsubsection{Performance against Sparse Data}
\begin{figure}[t]
\vspace{-0.1pt}
	\centering
	\begin{minipage}{0.32\linewidth}
		\centerline{\includegraphics[width=\textwidth]{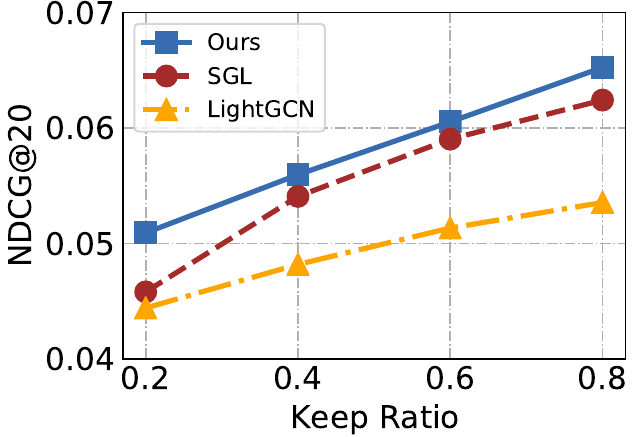}}
		\centerline{\;\;\;\;\;\;(a) Yelp2020}
	\end{minipage}
	\begin{minipage}{0.32\linewidth}
		\centerline{\includegraphics[width=\textwidth]{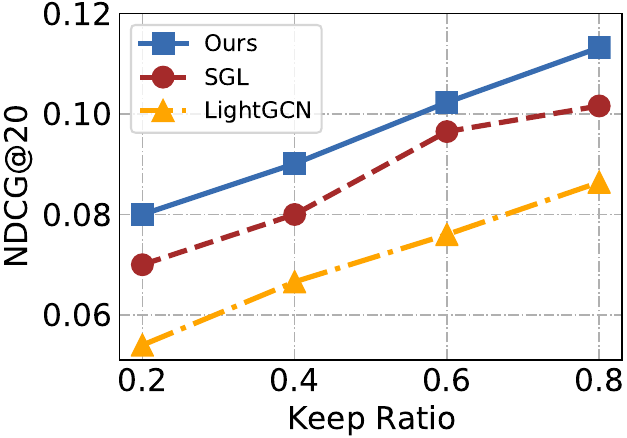}}
		\centerline{\;\;\;\;\;\;(b) Amazon-CD}
	\end{minipage}
	\begin{minipage}{0.32\linewidth}
		\centerline{\includegraphics[width=\textwidth]{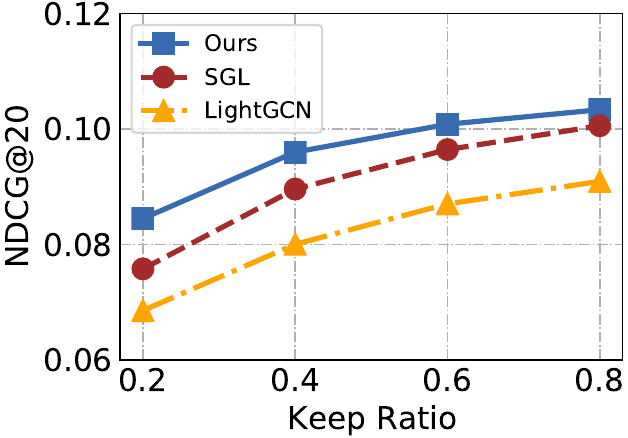}}
		\centerline{\;\;\;\;\;\;(c) Amazon-Book}
	\end{minipage}
	\caption{Performance on the Yelp2020 and Amazon-Book datasets with varying degrees of sparsity level.}
	\label{fig4_1}
\end{figure}

We investigate the robustness of the model against varying levels of sparsity in different proportions of the training datasets. Specifically, we remove a certain percentage of interaction data and then test SGCL and baseline models using training sets with different ratios (ranging from 20\% to 80\%). As shown in Figure \ref{fig4_1}, the performance of SGCL across different proportions of training datasets consistently outperforms baseline methods. This suggests the robust capability of SCL against noisy views.
Furthermore, as evident from the graph, the performance of self-supervised methods (e.g., SGCL and SGL) significantly outperforms non-self-supervised methods (e.g., LightGCN). This further substantiates the notion that self-supervised methods generate more meaningful supervisory signals in sparse training sets, thereby mitigating to some extent the challenges associated with suboptimal recommendation performance due to data sparsity.

\begin{figure}[t]
	 \centering
	\begin{minipage}{0.32\linewidth}
		\vspace{5pt}
    \centerline{\includegraphics[width=1\textwidth]{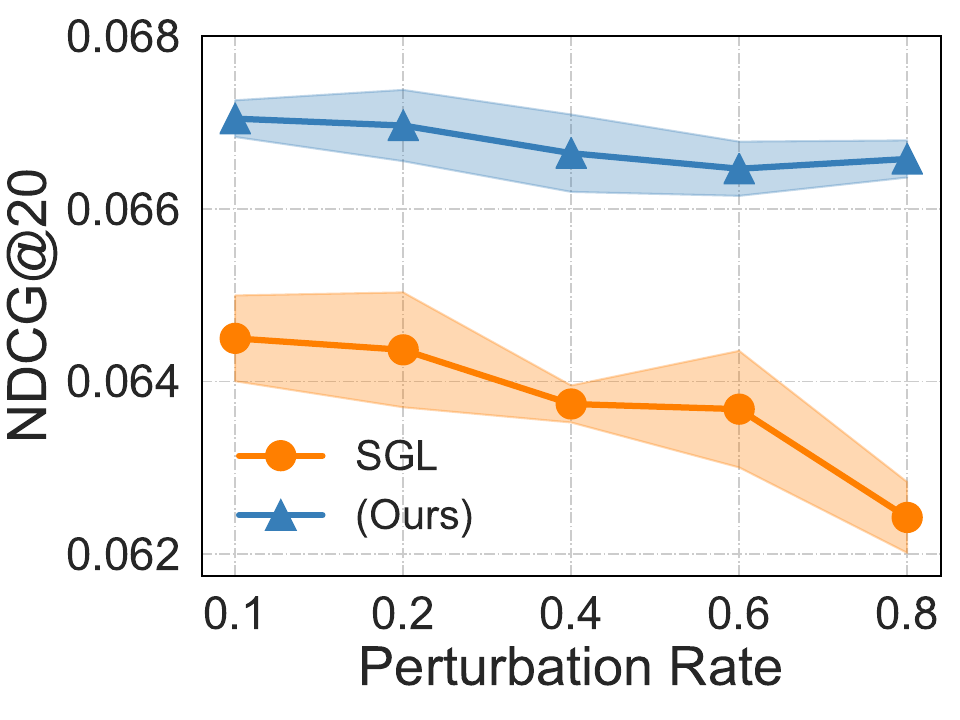}}
		\centerline{\;\;\;\;\;\;\;\;\;(a) Yelp2020}
	\end{minipage}
	\begin{minipage}{0.32\linewidth}
		\vspace{5pt}
\centerline{\includegraphics[width=1\textwidth]{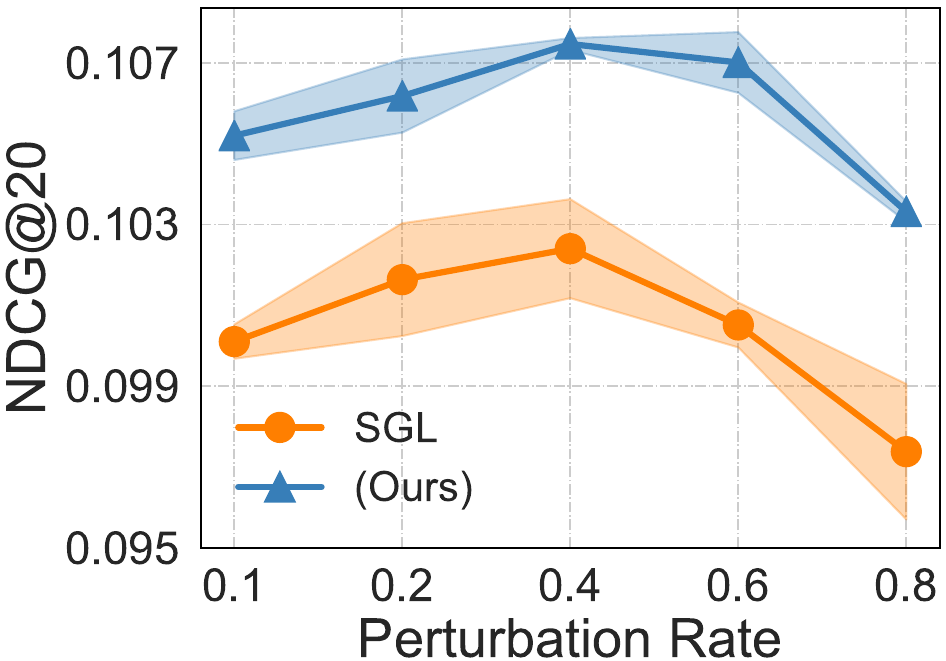}}
		\centerline{\;\;\;\;\;\;\;\;(b) Amazon-CD}
	\end{minipage}
 \begin{minipage}{0.32\linewidth}
		\vspace{5pt}
\centerline{\includegraphics[width=1\textwidth]{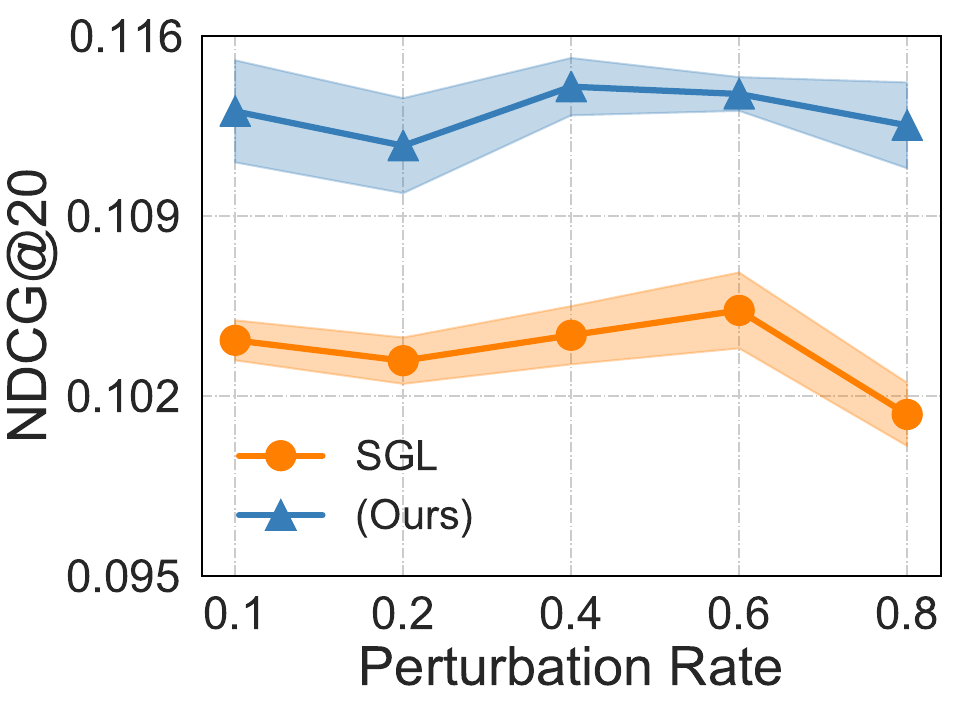}}
		\centerline{\;\;\;\;\;\;\;\;(c) Amazon-Book}
	\end{minipage}
	\caption{Performance v.s. Perturbation Rate. We increased the perturbation rate of edge dropout from 5\% to 25\%. }
	\label{fig5}
\end{figure}

\begin{table}[t]
\centering
 \caption{Performance comparison on Amazon-CD and Yelp2020 with different GNN backbone.}
 \renewcommand{\arraystretch}{1} 
 \setlength{\tabcolsep}{0.8mm}{ 
    \begin{tabular}{ccccccc}
     \hline
     \rowcolor{gray!20}  \multirow{2}{*}{Method}& \multicolumn{3}{c}{Amazon-CD}  & \multicolumn{3}{c}{Yelp2020}\\
        & Precision@20 & Recall@20 & NDCG@20 & Precision@20 & Recall@20 & NDCG@20 \\
    \hline
    NCL & 0.0275 & 0.1464 & 0.0922 & 0.0247& 0.0912 & 0.0621 \\
    HCCF & 0.0290 & 0.1502 & 0.0970 & 0.0242 & 0.0900 &  0.0620 \\
    SGL-ND  & 0.0277 & 0.1464 & 0.0940 & 0.0210 &0.0884 & 0.0650 \\
    SGL-RW  & 0.0300 & 0.1590 & 0.1024 & 0.0220 &0.0880 & 0.0600 \\
    \hline
    NCL w/ SCL & \textbf{0.0290} &\textbf{0.1575} &  \textbf{0.0939} & \textbf{0.0265}& \textbf{0.0938} &  \textbf{0.0653}\\
  HCCF w/ SCL & \textbf{0.0305 } & \textbf{0.1685} & \textbf{0.1050} & \textbf{0.0263} & \textbf{0.0974} & \textbf{0.0632}  \\
   SGL-ND w/ SCL & \textbf{0.0320} & \textbf{0.1644} & \textbf{0.1084} & \textbf{0.0252} & \textbf{0.0911} & \textbf{0.0661}\\
   SGL-RW w/ SCL & \textbf{0.0323} & \textbf{0.1683}  & \textbf{0.1104}  & \textbf{0.0275} & \textbf{0.0920} & \textbf{0.0670}\\
    \hline
    \end{tabular}}
    \label{tab: 07}
\end{table}

\begin{table}[t]
\centering
 \caption{Performance comparison with different noise ratios on Amazon-CD, Amazon-Book, and Yelp2020 datasets.}
 \renewcommand{\arraystretch}{1} 
 \setlength{\tabcolsep}{1.5mm}{ 
    \begin{tabular}{ccccccccl}
     \hline
         \rowcolor{gray!20}  Datasets & Method & 0.1 & 0.15 &0.2 & 0.25 & 0.4 \\
    \hline
    \multirow{4}{*}{Amazon-CD}& NCL & 0.0836 & 0.0686 &0.0600  & 0.0548 & 0.0360 \\
    & \textbf{NCL w/ SCL} & \textbf{0.0882} & \textbf{0.0802}&\textbf{0.0736} &  \textbf{0.0690} & \textbf{0.0523} \\
    & HCCF & 0.0795 & 0.0725 &  0.0629 & 0.0560 & 0.0498 \\
    & \textbf{HCCF w/ SCL} & \textbf{0.0802} & \textbf{0.0783} & \textbf{0.0640} & \textbf{0.0588} & \textbf{0.0540}\\
    \hline
    \multirow{4}{*}{Amazon-Book}& NCL & 0.1043 & 0.0902 & 0.0812  & 0.0624 & 0.0536 \\
    & \textbf{NCL w/ SCL} & \textbf{0.1110} & \textbf{0.1020}&\textbf{0.0836} &  \textbf{0.0690} & \textbf{0.0605} \\
    & HCCF & 0.1099 & 0.0925 &  0.0729 & 0.0701 & 0.0608 \\
    & \textbf{HCCF w/ SCL} & \textbf{0.1207} & \textbf{0.1183} & \textbf{0.1040} & \textbf{0.0988} & \textbf{0.0704}\\
    \hline
    \multirow{4}{*}{Yelp2020}& NCL & 0.0601 & 0.0556 &0.0500  & 0.0448 & 0.0260 \\
    &\textbf{NCL w/ SCL} & \textbf{0.0659} & \textbf{0.0606}&\textbf{0.0536} &  \textbf{0.0490} & \textbf{0.0323} \\
    & HCCF & 0.0610 & 0.0525 &  0.0429 & 0.0480 & 0.0398 \\
    & \textbf{HCCF w/ SCL} & \textbf{0.0682} & \textbf{0.0640} & \textbf{0.0540} & \textbf{0.0500} & \textbf{0.0440}\\
    \hline
    \end{tabular}}
    \label{tab: 08}
\end{table}

\begin{figure}[ht]
\setlength{\belowcaptionskip}{-0.5cm} 
	 \centering
	\begin{minipage}{0.8\linewidth}
		\vspace{2pt}
    \centerline{\includegraphics[width=\textwidth]{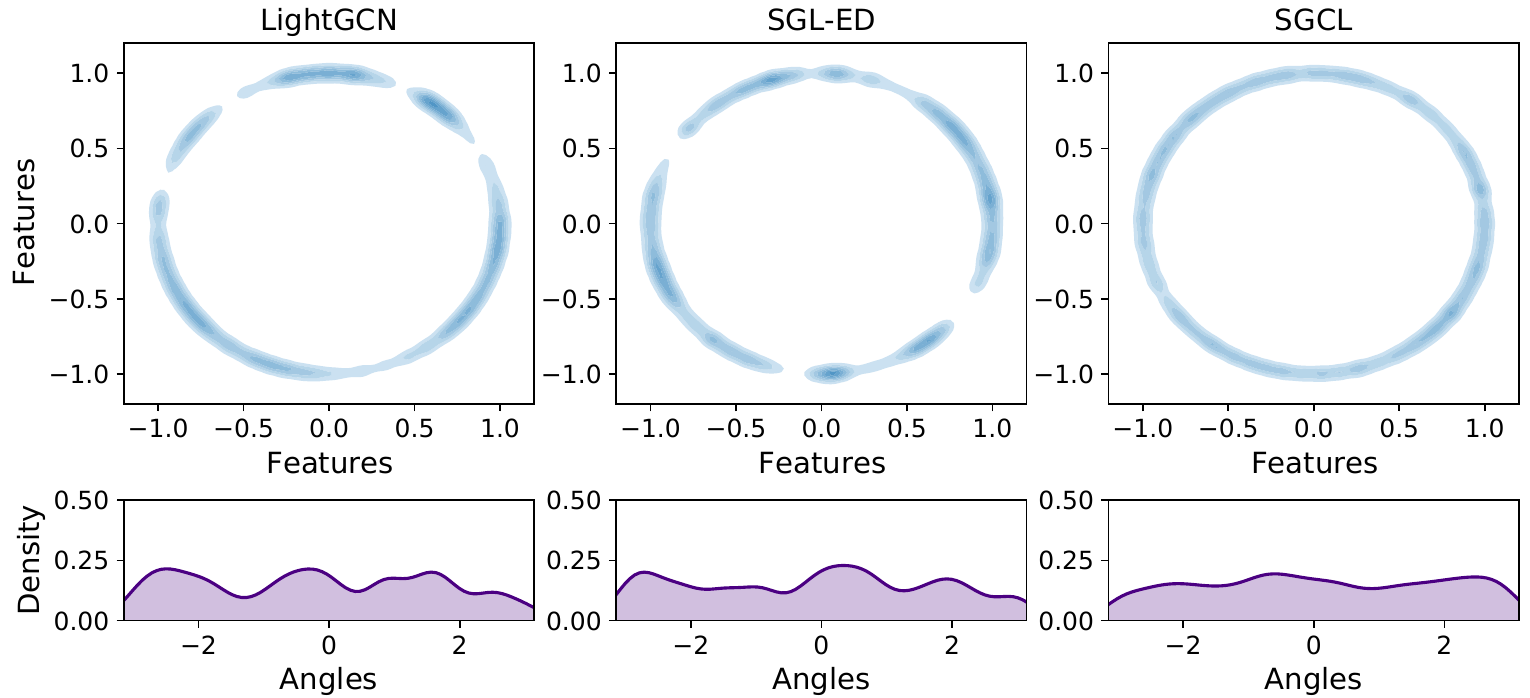}}
	\end{minipage}

 
	\caption{User embedding distribution visualization of different methods on Yelp2020 dataset (the darker the color is, the more
points fall in that area). }
	\label{fig6}
\end{figure}

\subsection{Ablation Study (RQ3)}
We carry out ablation experiments to analyze the effect of SCL within SGCL on model performance, with results presented in Figure \ref{fig5}. Additionally, we extend SGL to other GNN-based architectures and report the outcomes in Table \ref{tab: 07} and Table \ref{tab: 08}.

Figure \ref{fig5} illustrates the influence of various perturbation rates on model performance across the Yelp2020, Amazon-Book, and Amazon-CD datasets. Specifically, we escalate the perturbation rate for edge perturbation from 10\% to 80\%. The figures show that with the increase in perturbation rate, SGCL consistently outperforms SGL on both datasets, with SGCL also exhibiting a minor variance. This further validates the robustness of SCL against noisy views and its capability to enhance model performance. Simultaneously, it can be observed that the model's sensitivity to perturbation rate varies across different datasets. For example, the SGL's performance on Amazon-CD deteriorates significantly as the perturbation rate increases from 60\% to 80\%.
We attribute this to the denser nature of the Amazon-CD dataset. As the perturbation rate increases, disruptions in the original graph connections introduce noisy views, leading to suboptimal user/item representations and ultimately impacting the model's performance.

In Table \ref{tab: 07},
We integrate SCL into another GNN-based contrastive learning method and compare them with their original methods on the Amazon-CD dataset and Yelp2020 dataset. SGL-ED represents the SGL model with the node dropout data augmentation method, while SGL-WK uses the random walk. From the table \ref{tab: 07}, we can see that using the SCL contrastive loss function results in performance improvements compared to their original models. For example, the SGL-RW (W/ SCL) metrics improved by 5.85\%  to 7.81\% on the Amazon-CD dataset compared with the original SGL-RW methods, and the SGL-ND (w/ SCL) metrics improved by  12.30\% to 15.52\% on the Amazon-CD dataset compared with the original SGL-ND methods. {Therefore, we can draw an important conclusion: Our proposed noise-tolerant contrastive learning method, SCL, effectively manages noise introduced by various manual data augmentation techniques, thus enhancing the accuracy of our recommendations.}. Furthermore, the performance of NCL (with SCL) and HCCF (with SCL) improved on both the Amazon-CD and Yelp2020 datasets. This suggests that the SCL method can also boost recommendation performance when integrated with other GNN-based recommendation methods that employ adaptive or embedding augmentation. Table \ref{tab: 08} reports the performance of models across different noise ratios on the Amazon-CD, Amazon-Book, and Yelp2020 datasets. As the noise ratio in the training data increases, it becomes apparent that the performance of the models declines. However, even when the noise ratio reaches 40\%, NCL (w/ SCL) still shows an average improvement of 27\% compared with NCL, and HCCF (w/ SCL) shows an average improvement of 10\%. The notable enhancement in the performance of NCL (with SCL) and HCCF (with SCL) further validate the efficacy of our proposed approach.

\ \\\noindent
\textbf{Embedding visualisation analysis}.
We randomly select 3000 user nodes from the Yelp2020 and Amazon-CD datasets and replace 25\% genuine interactions with fake interactions. Subsequently, we utilize t-SNE\cite{cai2022theoretical} for embedding visualization, projecting them into a 2-D space. Figure \ref{fig6} shows that the features learned by SGCL exhibit a more uniform distribution than the other two methods. The representations learned by all contrastive learning methods are more uniformly distributed than the supervised method. The density estimation curves of LightGCL are sharp in both datasets. Previous research \cite{yu2022graph,lin2022improving} corroborates the idea that the uniformity of distribution significantly influences recommendation performance. Therefore, the findings depicted in Figure \ref{fig6} provide additional validation of the resilience of SCL against noisy views and its efficacy in enhancing recommendation accuracy.

\subsection{Hyperparameter Investigation (RQ4)}

\begin{figure}[t]
	 \centering
	\begin{minipage}{0.45\linewidth}
		\vspace{5pt}
    \centerline{\includegraphics[width=1\textwidth]{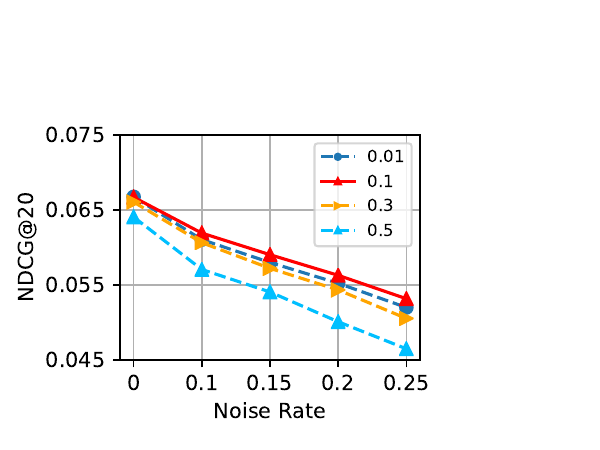}}
		
	\end{minipage}
	\begin{minipage}{0.45\linewidth}
		\vspace{5pt}
\centerline{\includegraphics[width=1\textwidth]{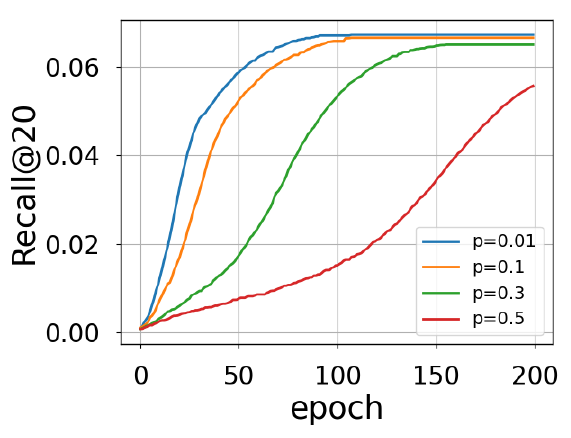}}

	\end{minipage}

 \caption{The experiments on the Amazon-CD dataset aim to explore the impact of the $p$ value on the model performance. The left figure illustrates the impact of different noisy rates on model performance at various values of $p$. On the right, the convergence speed of the model is depicted under different values of $p$.}
	\label{fig: 07}
\end{figure}

\begin{figure}[t]
	\centering
	\begin{minipage}{1\linewidth}
		\vspace{3pt}
	\centerline{\includegraphics[width=\textwidth]{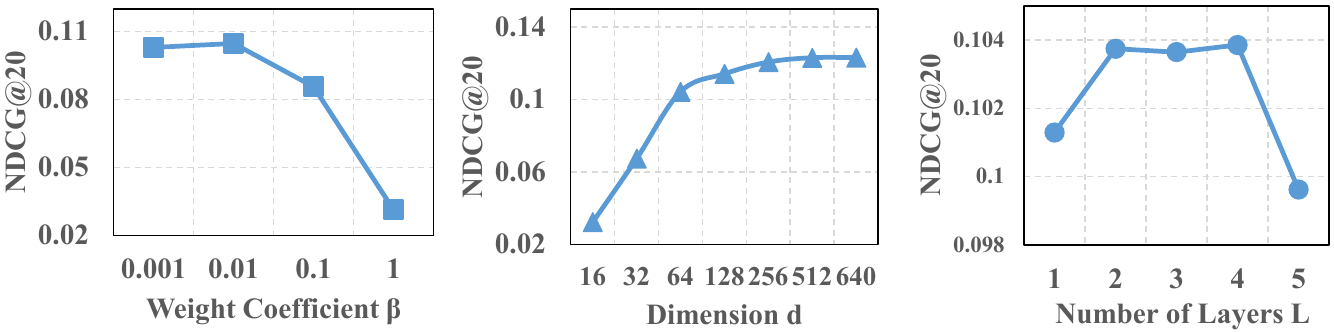}}
    \centerline{\;\;\;\;\;(a) Amazon-CD}
			\vspace{1pt}
		\centerline{\includegraphics[width=\textwidth]{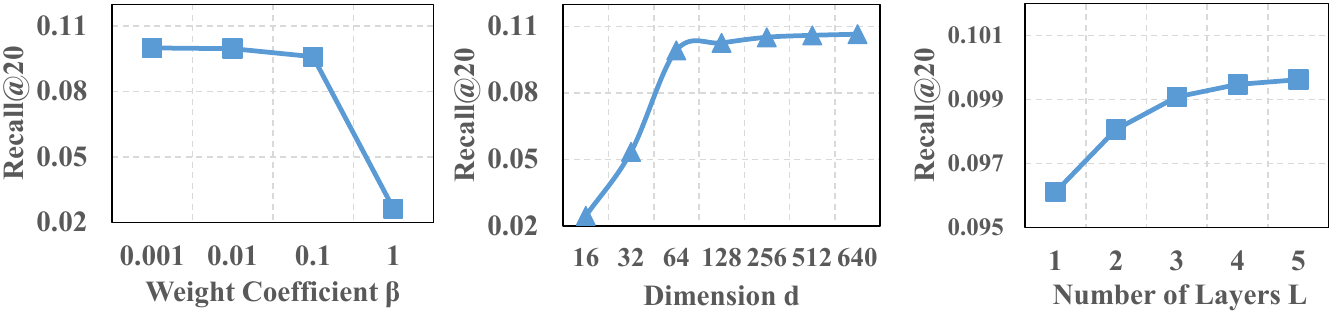}}
    \centerline{\;\;\;\;\;(b) Yelp2020}
			\vspace{1pt}	
	\end{minipage}
	\caption{The outcomes with various hyperparameter configurations concerning Recall@20 and NDCG@20 on both the Amazon-CD and Yelp2020 datasets.}
	\label{fig8}
\end{figure}

\noindent
\textbf{Convergence and robustness trade-off coefficient $p$}: We evaluate the effect of different values of $p$ on the model's performance using the Yelp2020 dataset. The left section of Figure \ref{fig: 07} demonstrates how varying values of $p$ affect model performance across different levels of noise. The graph indicates that the model performs comparably well when $p$ is set to 0.01 and 0.1, showing better recommendation accuracy than at other values of $p$. Additionally, as the noise rate escalates, the decline in model performance is less pronounced at $p$ values of 0.01 and 0.1. In the right section of Figure \ref{fig: 07}, we document the model's convergence with various settings of $p$. The convergence speed of the model decreases as the value of $p$ increases. Thus, to strike a balance between the model's robustness and convergence speed in practical recommendation scenarios, it is advisable to select a $p$ value within the range of [0.01, 0.1].

\ \\\noindent
\textbf{Weight coefficient $\beta$ in loss function}: This hyperparameter controls the regularization strength of $\mathcal{L}_{SCL}$, which is an auxiliary task to help the representation learning. Observations from the left section of Figure \ref{fig8} reveal that setting $\beta$ to 0.001 and 0.01 markedly enhances the model's performance. Conversely, applying larger $\beta$ values leads to a sharp deterioration in performance across both datasets. We argue that the reason for the performance decline is overfitting. Therefore, in the training process of the SGCL, a common choice for $\beta$ is 0.01.

\ \\\noindent
\textbf{Dimension of latent space $d$}:
In the middle section of Figure \ref{fig8}, we document the effects of varying embedding sizes, ranging from 16 to 640, on the performance of the model on the Amazon-CD and Yelp2020 datasets. Initially, an increase in the embedding size substantially boosts the model's performance, as larger embeddings can encapsulate more complex language details, resulting in improved representations of users (items). However, beyond an embedding size of 256, further enlargements do not enhance performance and instead lead to increased computational demands.
    
\ \\\noindent
\textbf{Number of layers $L$}:  
The right side of Figure \ref{fig8} explores the influence of varying numbers of layers on model performance across diverse datasets. It is apparent that as the number of layers increases, so does the model's performance. However, on the Amazon-CD dataset, performance begins to decline when the number of layers reaches 5, suggesting potential overfitting. Generally, setting the number of convolutional layers to 4 yields optimal recommendation performance."
\section{Related Work}

\textbf{Graph Neural Network for Collaborative Filtering}.  Recently, graph models have been proposed for representation learning, such as GCN \cite{abadal2021computing, gao2023survey, wu2020comprehensive}, GraphSAGE \cite{hamilton2017inductive, liu2020graphsage}, GAT \cite{velickovic2017graph, brody2021attentive}, HetGNN \cite{zhang2019heterogeneous,li2021higher}, and HGNN \cite{feng2019hypergraph} etc.
These techniques are extensively applied across multiple graph tasks in computer vision (CV) and natural language processing (NLP), including node classification, graph classification, and link prediction.
Thanks to the robust representation learning capabilities of graph models, user-item interaction data is inherently structured into what is known as the user-item interaction graph. Graph Neural Networks (GNNs) leverage this structure to learn about user preferences. Various GNN-based collaborative filtering approaches have been developed by researchers \cite{wu2022graph, gao2023survey}. Commonly, these approaches utilize a GNN to generate embeddings that represent users (or items), followed by the use of inner product operations to predict the probability of user-item interactions. For instance, NGCF \cite{wang2019neural}, a well-known graph neural collaborative filtering technique, constructs embedding propagation layers to capture high-order connections within the user-item interaction graph, resulting in notable performance gains. Recent methods like LightGCN\cite{he2020lightgcn} and GCCF \cite{chen2020revisiting} eliminate non-linear transformations and activations during embedding propagation to streamline the graph message passing algorithm. DGCF \cite{wang2020disentangled} designs a disentanglement module for learning diverse user intents and modeling the diversity of relationships between users and items. Current approaches suggest that models grounded in Euclidean space might not entirely grasp the complexities of user-item interactions. Consequently, research is exploring graph collaborative filtering methods that utilize hyperbolic space. For example, GDCF \cite{zhang2022geometric} aims to enhance recommendation accuracy by learning geometric disentangled representations to uncover latent intent factors across multiple geometric spaces. HNCR \cite{li2022hyperbolic} attempts to enhance recommendation performance by leveraging hyperbolic geometry and deep learning techniques. Knowledge graphs are integrated with graph neural networks to improve recommendation accuracy and enhance result interpretability \cite{peng2023knowledge}. For example, KGAT \cite{wang2019kgat}, KPRN \cite{wang2019explainable}, and KGCN \cite{wang2019knowledge}. Despite the success of these methods, their model performance relies on high-quality supervision signals. Recommendation performance is constrained when facing sparse and noisy data.

\ \\\noindent
\textbf{Self-supervised Graph Learning for  Recommendation}. Self-supervised learning (SSL) \cite{jaiswal2020survey} can alleviate label scarcity and noisy issues in recommendation systems by generating self-supervised signals. Graph contrastive learning with data augmentation is a standard recommendation paradigm. Existing studies design various graph augmentation schemes for embedding contrasts. For example, SGL\cite{wu2021self} employs random node/edge dropout to generate contrastive views. However, some works argue that such random data augmentation will break the original connection, generate noisy views, and result in suboptimal representation\cite{yu2022graph}. Therefore, researchers have proposed some adaptive contrastive learning frameworks. For instance, AdaMCL \cite{zhu2023adamcl} introduces an adaptive multi-view fusion contrastive learning framework. GFormer \cite{Li2023gformer} devises an adaptive framework by combining generative self-supervised learning with graph transformer architecture for rationale discovery. CGCL \cite{he2023candidate} investigates the correlation between users and candidate items and utilizes comparable semantic embeddings to form contrasting pairs. VGCL \cite{yang2023generative} learns the latent distribution of nodes through variational inference and generates contrastive views through multiple samplings.
Although those above adaptive contrastive learning frameworks can avoid generating noisy views and boost recommendation performance, their network structures are typically more complex. This paper introduces a simple and general SGCL framework from the perspective of loss functions, which effectively enhances model robustness and improves recommendation accuracy.

\section{Conclusions and Future Work}

In this study, we initially conduct experiments to examine the potential noise generated by directly applying manual data augmentation to the user-item graph. To be specific, we define noisy views as the final 20\% of generated views with cosine similarity to the original graph below 0.1. Our experiments demonstrate that the number of noise views generated increases randomly as key nodes and edges are removed. Furthermore, the experimental results further show that these noisy views directly compromise the accuracy of the recommendation system.
Therefore, this work proposes a theoretically guaranteed robust recommender system, SGCL, from the perspective of the loss function. In our SGCL, we suggest the symmetric contrastive loss and provide theoretical proof demonstrating the robustness of SCL in the presence of noisy views. 
Our experiments conducted on three real-world datasets indicate that SGCL surpasses the state-of-the-art baselines and notably enhances recommendation performance. The analysis on robustness confirms the resilience of our SGCL model in handling noise and sparse data.

An intriguing direction for future research involves the application of causal inference techniques to generate self-supervised signals. By examining the causal interactions between users and items from a causal viewpoint, we can produce high-quality self-supervised signals that improve the interpretability of the model’s recommendations.


\section*{ACKNOWLEDGMENTS}
This work is partially supported by the National Natural Science Foundation of China under Grant No. 62032013, the Science and technology projects in Liaoning Province (No. 2023JH3/10200005), and the Fundamental Research Funds for the Central Universities under Grant No. N2317002.

\bibliographystyle{ACM-Reference-Format}
\bibliography{sample-base}


\end{document}